# MATHEMATICAL MODELS IN SCHEMA THEORY

**Mark Burgin**

Department of Mathematics
University of California, Los Angeles
405 Hilgard Ave.
Los Angeles, CA 90095

**Abstract:** In this paper, a mathematical schema theory is developed. This theory has three roots: brain theory schemas, grid automata, and block-shemas. In Section 2 of this paper, elements of the theory of grid automata necessary for the mathematical schema theory are presented. In Section 3, elements of brain theory necessary for the mathematical schema theory are presented. In Section 4, other types of schemas are considered. In Section 5, the mathematical schema theory is developed. The achieved level of schema representation allows one to model by mathematical tools virtually any type of schemas considered before, including schemas in neurophisiology, psychology, computer science, Internet technology, databases, logic, and mathematics.

**Keywords:** schema, grid automaton, schema multigraph, schema homomorphism, concretization

## 1. Introduction

Schema theory is an approach to knowledge representation, organization, processing, and utilization. The concept of schema is extensively used in psychology, theory of learning, and the theory of programming. We analyze the version called interaction schema that has been explicitly shaped by the need to understand how cognitive and instinctive functions can be implemented in a distributed fashion such as that involving the interaction of a multitude of brain regions (Arbib, 1985; 1989; 1992; 1995). Many of the concepts have been abstracted from biology to serve as "bridging" concepts which can be used in both for the study of interacting agents in AI and brain theory and thus, can serve cognitive science whether or not the particular study addresses neurological or neurophysiological data. Our aim of this paper is to develop mathematical foundations for schema theory, building a mathematical schema theory.

As the base for the development of this mathematical theory, we take the construction of a grid automaton (Burgin, 2003a; 2003b; 2005). Examples of grid automata are numerous: neural networks, cellular automata, Petri nets, random access machines (RAM), Turing machines, finite automata, port automata, state machines, and inductive Turing machines are all grid automata. At the same time, grid automata have their advantages in comparison with all these constructions. In comparison with cellular automata, a grid automaton can contain different kinds of automata as its nodes. For example, finite automata, Turing machines and inductive Turing machines can belong to one and the same grid. In comparison with systolic arrays, connections between different nodes in a grid automaton can be arbitrary like connections in neural networks. In comparison with neural networks and Petri nets, a grid automaton may contain, as its nodes, machines more powerful than finite automata. Consequently, neural networks, cellular automata, systolic arrays, and Petri nets are special kinds of grid automata. An important property of grid automata is the possibility of realizing hierarchical structures, that is, a node can itself be a grid automaton. In grid automata, interaction and communication becomes as important as computation. This peculiarity results in a variety of types of automata, their functioning modes, and space organization. All these properties make grid automata a suitable frame for the development of a mathematical schema theory. In addition, specific grid automata, e.g., neural networks, are instantiations and realizations of schemas.

Here we introduce and study a general mathematical concept of schema on a level of generality that makes it possible to model by mathematical tools virtually any type of schemas considered before, including schemas in neurophisiology, psychology, computer science, Internet technology, databases, logic, and mathematics. The reason for such high level development is existence of different types of schemas (in brain theory, cognitive psychology, artificial intelligence, programming, computer science, mathematics, databases, etc.). To better understand human intellectual activity (thinking, decision-making, and learning) and to build artificial intelligence, we have to be able to work with a variety of schema types. In our analysis, we have to go from large information processing blocks down to the finest details of neural structure and function.

Our next step will be in developing a specialization of the general mathematical schema theory oriented at neurophisiological understanding of human thinking and action. At this point, interaction schemas become a specialization of the general theory. Formalization of interaction schemas enables us to define and study relations between schemas and operations with schemas in a more exact and efficient way.

It is necessary to remark that definitions introduced in this paper are not purely formal constructions but represent mathematical models for a variety of important real phenomena and systems. Providing such a generality, these definitions serve our main goal here – the development of mathematical means for neurophisiological studies. For instance, concepts of schema determination, concretization, abstraction, and homomorphism give mathematical models for basic operations and actions in thinking, decision-making, and learning.

As Deloup writes (2005), "understanding why one definition rather than another is "right" is a fine art, and there is much room for argument about it. However, this kind of understanding lies at the core of doing mathematics." One of the most noteworthy examples for this claim is Alan Turing. Now he is, may be, the most famous computer scientist and all know him not because of his theorems but due to his invention of the most popular model of computation – Turing machine.

The author is grateful to Michael Arbib for fruitful discussions and useful remarks.

## 2. Elements of the grid automaton theory

Informally, a grid automaton is a system of automata, which are situated in a grid, optionally connected, and interact with one another through a system of connections/links. The basic idea of interacting devices and communicating automata and processes is for a transmitting system/process to send a message to a port and for receiving system/process to get the message from a port. Thus, to formalize this structure, we assume, as it is often true in reality, that connections are attached to automata by means of ports. Ports are specific automaton elements through which information/data come into (*output ports* or *outlets*) and send outside the automaton (*input ports* or *inlets*). Thus, any system $P$ of ports is the union of its two disjunctive, i.e., nonintersecting, subsets $P = P_{in} \cup P_{out}$ where $P_{in}$ consists of all inlets from $P$ and $P_{out}$ consists of all outlets from $P$. If there are ports that are both inlets and outlets, we combine such ports from couples of an input port and an output port.

There are different other types of ports. For example, contemporary computers have parallel and serial ports. Ports can have inner structure, but in the first approximation, it is possible to consider them as elementary units.

We also assume that each connection is directed, i.e., it has the beginning and end. It is possible to build bidirectional connections from directed connections.

**Definition 2.1.** A *grid automaton G* is the following system that consists of three sets and three mappings

$$G = (A_G, P_G, C_G, p_{IG}, c_G, p_{EG})$$

Here:

The set $A_G$ is the set of all automata from $G$;

the set $C_G$ is the set of all connections/links from $G$;

the set $P_G = P_{IG} \cup P_{EG}$ (with $P_{IG} \cap P_{EG} = \varnothing$) is the set of all ports of $G$, $P_{IG}$ is the set of all ports (called *internal ports*) of the automata from $A_G$, and $P_{EG}$ is the set of *external ports* of $G$, which are used for interaction of $G$ with different external systems;

$p_{IG} : P_{IG} \to A_G$ is a total function, called the *internal port assignment function*, that assigns ports to automata;

$c_G : C_G \to (P_{IGout} \times P_{IGin}) \cup P'_{IGin} \cup P''_{IGout}$ is a (eventually, partial) function, called the *port-link adjacency function*, that assigns connections to ports where $P'_{IGin}$ and $P''_{IGout}$ are disjunctive copies of $P_{IGin}$.

and

$p_{EG} : \boldsymbol{P}_{EG} \to \boldsymbol{A}_G \cup \boldsymbol{P}_{IG} \cup \boldsymbol{C}_G$ is a function, called the *external port assignment function*, that assigns ports to different elements from $G$.

If $l$ is a link that belongs to the set $\boldsymbol{C}_G$ and $c_G(l)$ belongs to $\boldsymbol{P}_{Gin} \times \boldsymbol{P}_{Gout}$, i.e., $c_G(l) = (p_1, p_2)$, it means that the beginning of $l$ is attached to $p_1$, while the end of $l$ is attached to $p_2$. Such link is called *closed*. If $l$ is a link from $\boldsymbol{C}_G$ and $c_G(l)$ belongs to $\boldsymbol{P}_{Gin}$ (or $\boldsymbol{P}_{Gout}$), i.e., $c_G(l) = p_1 \in \boldsymbol{P}_{Gin}$ (correspondingly, $c_A(l) = p_2 \in \boldsymbol{P}_{Gout}$), it means that the beginning of $l$ is attached to $p_1$ (correspondingly, the end of $l$ is attached to $p_2$). Such links is called *open*.

The automata from $\boldsymbol{A}_G$ are also called *nodes of G*, and connections/links from $\boldsymbol{C}_G$ are also called *edges of G*. Like ports, nodes and edges can be of different types. For instance, nodes in a grid automaton can be neurons, neural networks, finite automata, Turing machines, port automata (Arbib, Steenstrup, and Manes, 1983), vector machines, array machines, random access machines (RAM), inductive Turing machines (Burgin, 2005), fuzzy Turing machines (Wiedermann, 2004), etc. Even more, some of the nodes can be also grid automata.

As a result, elements from the set $\boldsymbol{A}_G$ have they inner structure. Besides, elements from the sets $\boldsymbol{P}_G$ and $\boldsymbol{C}_A$ can also have they inner structure. For example, a link or a port can be an automaton. If we consider Internet as a grid automaton with computers as nodes, then links include modems, servers, routers, and eventually some other devices. A network adapter is an example of a port with inner structure.

**Remark 2.1.** To have meaningful assignments of ports, the port assignment functions $p_{IG}$ and $p_{EG}$ have to satisfy some additional conditions. For instance, it is necessary not to assign (attach) input ports of the automaton $G$ to the end (output side) of any link in $G$. In the case of a neural network as a node of $G$, inner ports of $G$ are assigned to this network are usually connected to open links going to (inlets) and from (outlets) neurons. At the same time, it is possible to have such ports connected to neurons directly, as well as free ports that are not connected to any element of the network. Free ports might be useful for increasing reliability of the network connections to the environment. When some port fails, it would be possible by dynamically changing the assignment function to change the damaged port by a free port.

Taking the nervous system of a human being and representing it as a grid automaton with neurons as its nodes, it natural to consider dendrites and axons as links – dendrites are input (incoming) links and axons are output (outgoing) links. Then synaptic membranes are ports of this automaton: presynaptic membranes are outlets and postsynaptic membranes are inlets. Presynaptic membranes are axon terminals, i.e., output ports are adjusted only to output links, while postsynaptic membranes are parts of dendrites and bodies of neurons, i.e., input ports are adjusted both to nodes (automata) and to input links.

Cell membranes in general and neuron membranes, in particular, give examples of ports with a complex inner structure.

**Remark 2.2.** Representation of grid automata without ports is the first approximation to a general network model (Burgin, 2003), while representation of grid automata with ports is the second (more exact) approximation. In some cases, it is sufficient to use grid automata without ports, while in other situations to build an adequate, flexible and efficient model, we need automata with ports as nodes of a grid automaton.

To achieve better comprehensibility, grid automata are usually represented in a graphical form as in Figure 2.1.

A formal description of a grid automaton without ports is given in the following definition.

**Definition 2.2.** A *basic grid automaton A* is the following system that consists of two sets and one mapping

$R = (A_A, C_A, c_A)$

Here:

The set $A_A$ is the set of all automata from $A$;

the set $C_R$ is the set of all connections/links from $R$;

and

$c_A: C_A \rightarrow A_A \times A_A \cup A'_A \cup A''_A$ is a (variable) function, called the *node-link adjacency function*, that assigns connections to nodes where $A'_A$ and $A''_A$ are disjunctive copies of $A_A$.

**Example 2.1.** A grid automaton with such nodes as Turing machines, random access machines, neural networks, finite automata, cellular automata, and other grid automata is given in Figure 2.1.

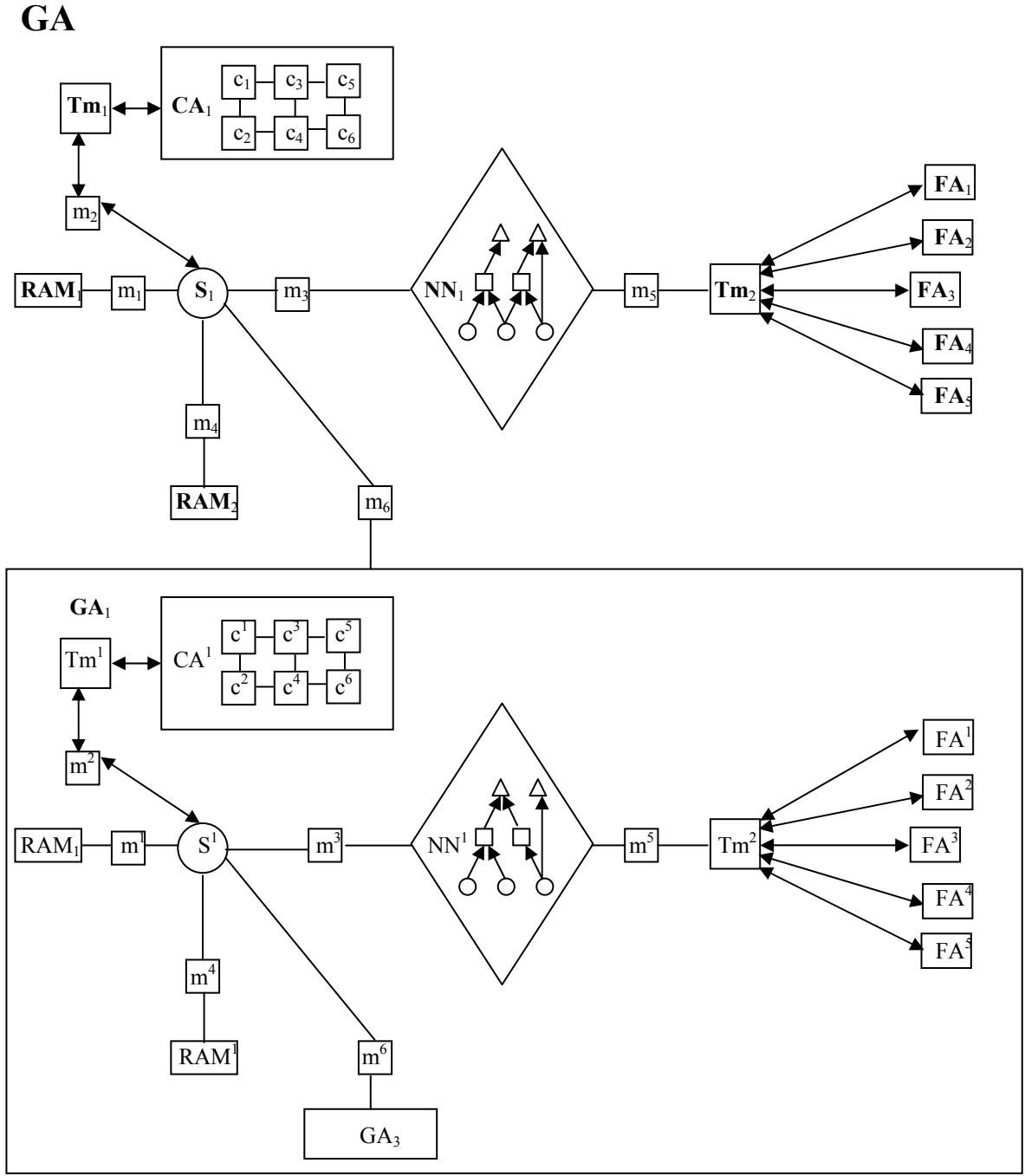

**Figure 2.1.** A grid automaton **GA** in which: **Tm** is a Turing machine; **RAM** is a random access machine; **S** is a server; **m** is a modem; **NN** is a neural network; **FA** is a finite automaton; and **CA** is a cellular automaton. Thus, this grid automaton **GA** contains as nodes: two Turing machines, one neural network, two RAM, five finite automata, one cellular automaton, six modems, one server, and one grid automaton.

Grid automata are abstract (mathematical) models of grid arrays. While a grid array consists of real/physical information processing systems and connections between them, a grid automaton consists of other abstract automata as its nodes and connections/links between them.

A grid automaton *B* is described by three grid characteristics, three node characteristics, and three edge characteristics.

<u>The grid characteristics are:</u>

1. The *space organization* or *structure* of the grid automaton *B*. This space structure may be in physical space, reflecting where the corresponding information processing systems (nodes) are situated, or it may be a mathematical structure defined by the geometry of node relations. Besides, we consider three levels of space structures: local, region, and global space structures of a schema. Sometimes these structures are the same, while in other cases they are different.

The space structure of a grid automaton can be *static* or *dynamic*. To get a more detailed classification, we assume that functioning of a grid constitutes of elementary operations, which can be discrete or continuous. In addition, these operations are organized so that they form definite cycles of computation and interaction. For instance, taking a finite automaton, we see that an elementary operation is processing of a single symbol, while a cycle is processing of a separate word. A cycle for a Turing machine is the process that goes from the start state to a final state of the machine. This gives us three kinds of space organization of a grid automaton: *static structure* that is always the same; *persistent dynamic structure* that may change between different cycles of computation; and *flexible dynamic structure* that may change at any time of computation. Persistent Turing machines (Goldin and Wegner, 1988) have persistent dynamic structure, while reflexive Turing machines (Burgin, 1992) have flexible dynamic structure.

2. The *topology* of *B* is determined by the type of the node neighborhood. A neighborhood of a node is the set of those nodes with which this node directly interacts.

In a physical grid, these are often the nodes that are the closest to the node in question. For example, if each node has only two neighbors (right and left), this may define either linear or circular topology in *B*. When there are four nodes (upper, below, right, and left), the *B* may have a two-dimensional rectangular topology. It could be also a topology of cylinder, torus or Möbius band.

Topology of computer networks gives an example of the grid automaton topology (Heuring and Jordan, 1997).

There are three main types of grid automaton topology:

- A *uniform topology*, in which neighborhoods of all nodes of the grid automaton have the structure.

- A *regular topology*, in which the structure of different node neighborhoods is subjected to some regularity. For instance, the system neighborhoods can be invariant with respect to gauge transformations similar to gauge transformations in physics (cf., for example, (Yndurain, 1983).

- An *irregular topology* where there is no regularity in the structure of different node neighborhoods.

Conventional cellular automata have a uniform topology. Cellular automata in the hyperbolic plane or on a Fibonacci tree (Margenstern, 2002) give an example of grid automata with a regular topology.

3. The *dynamics* of *B* determines by what rules its nodes exchange information with each other and with the environment of *B*. For example, it is possible that there is an automaton *A* in *B* that determines when and how all automata in *B* interact. Then if the automaton *A* is equivalent to a Turing machine, i.e., *A* is a recursive algorithm (Burgin, 2005), and all other automata in the grid automaton *B* are also recursive, then *B* is equivalent to a Turing machine (Burgin, 2003). At the same time, when the interaction of Turing machines in a grid automaton *B* is random, then *B* is much more powerful than any Turing machine (Burgin, 2003).

Interaction with the environment separates two classes of grid automata: *open grid automata* interact with the environment through definite connections, while *closed grid automata* have no interaction with the environment. For example, Turing machines are usually considered as closed automata because they begin functioning from some start state and tape configuration, finish functioning (if at all) in some final state and tape configuration, and do not have any interactions with their environment.

In turn, there are three types of open grid automata:

1. Grid automata open only for reception of information from the environment. They are called *accepting grid automata* or acceptors.
2. Grid automata open only for sending their output to the environment. They are called *transmitting grid automata* or transmitters.
3. Grid automata open for both receiving information from and sending their output to the environment. They are called *transducing grid automata* or transducers.

To be open, a grid automaton must have a definite topology. For instance, to be an acceptor, a grid automaton must have open input edges.

Existence of free ports makes a closed grid automaton *potentially open* as it is possible to attach connections to these ports.

The node characteristics are:

1. The *structure* of the node, including structures of its ports. For example, one node structure determines a finite automaton, while another structure is a Turing machine. It is possible that nodes also have they inner structure. For instance, representing the structure of a natural neuron, we can treat dendrites as ports. In this case, ports have rather developed inner structure, which can be represented on different levels – from functional components to molecular and even atomic organization.

In particular, the structure of a node defines how ports are adjusted in the node. For instance, in the case of a neural network that is a node of the grid automaton $A$, inner ports of $A$ are usually connected to links going to and from neurons. At the same time, it is possible to have ports connected to neurons directly, as well as free ports that are not connected to any element of the network. Free ports might be useful for reliability of the network connections to the environment.

In the case, of a Turing machine $T$ that is a node of the grid automaton $A$, it is possible to connect inner ports of $A$ to some cells of the tapes from $T$ or to whole tapes. In the first case, external information coming to such input ports will be written in the adjusted cells, while output ports will allow to send to another node (automaton) the symbol written in the cells to which these ports are adjusted. In the second case, external information coming to such input ports will be distributed in the corresponding tape by some rule, while an output port will allow to another node (automaton) the word written in the tape to which this port is adjusted.

2. The *external dynamics* of the node determines interactions of this node. According to this characteristic, there are three types of nodes: *accepting nodes* that only accept or reject their input; *generating nodes* that only produce some input; and *transducing nodes* that both accept some input and produce some input. Note that nodes with the same external dynamics can work in grids with various dynamics. Primitive ports do not change node dynamics. However, compound ports are able to influence processes not only in the node to which they belong but also in the whole grid automaton. For instance, a compound port can itself be an automaton.

3. The *internal dynamics* of the node determines what processes go inside this node. For instance, the internal dynamics of a finite automaton is defined by its transition function, while the internal dynamics of a Turing machine is defined by its rules. Differences in internal dynamics of nodes are very important because, for example, a change in producing the output allows us to go from conventional Turing machines to much more powerful inductive Turing machines of the first order (Burgin, 2005).

The edge characteristics are:

1. The *external structure* of the edge. According to this characteristic, there are three types of edges: a *closed edge* both sides of which are connected to ports of the grid automaton; an *ingoing edge* in which only the end side is connected to a port of the grid automaton; and an *outgoing edge* in which only the beginning side is connected to a port of the grid automaton

2. Properties and the *internal structure* of the edge. According to the internal structure, there are three types of edges: a *simple channel* that only transmits data/information; a *channel with filtering* that separates a signal from noise; and a *channel with data correction*.

3. The *dynamics* of the edge determines edge functioning. For instance, two important dynamic characteristics of an edge are bandwidth as the number of bits (data units) per second transmitted on the edge and throughput as the measured performance of the edge.

Properties of links/edges separate all links into three standard classes:

1. *Information link/connection* is a channel for processed data transmission.

2. *Control link/connection* is a channel for instruction transmission.

3. *Process link/connection* realizes control transfer and determines how the process goes by initiation of an automaton in the grid by another automaton (other automata) in the grid.

Process links determine what to do, control links are used to instruct how to work, and information links supply automata with data in a process of grid automaton functioning.

**Example 2.2.** When a sequential composition of two finite automata $A$ and $B$ is built, these automata are connected by two links. One of them is an information link. Through this link, the result obtained/produced by the first automaton $A$ is transferred from the output port (open from the right edge) of $A$ to the input port (open from the left edge) of $B$. In addition, $A$ and $B$ are connected by a control link. When the automaton $A$ produces its result, it transfers control to the automaton $B$. However, this does not mean that $A$ stops functioning – it can immediately start a new cycle of its functioning.

It is essential to remark that in some situations there are no control links between the automata in the composition and both are synchronized by data transfer.

**Remark 2.3.** Initiation of an automaton in the grid by a signal that comes through a control link is usually regulated by some condition(s). Examples of such conditions are: (a) some automata in the grid have obtained their results; (b) the initiated automaton has enough data to start working; (c) the number (level) of initiating signals is above a prescribed threshold. This is an event-driven functioning, which is usually contrasted with operating on a time-scale.

**Example 2.3.** Neurons (in a variety, but not all, models) are initiated only when the combined effect of all their input signals is above the firing threshold. For a natural neuron, single excitatory postsynaptic potentials (EPSPs) have amplitudes in the range of one millivolt. The critical value for spike initiation is about 20 to 30 mV above the resting potential. In most neurons, four spikes are not sufficient to trigger an action potential. Instead, about 20-50 presynaptic spikes have to arrive within a short time window before postsynaptic action potentials are triggered.

**Remark 2.4.** Transmission of instructions from one automaton in the grid to another one can be realized by transmission of values of some control parameter.

To represent structures of grid automata now and schemas later, we use oriented multigraphs and generalized oriented multigraphs.

**Definition 2.3** (Berge, 1973)**.** An *oriented* or *directed multigraph G* has the following form:

$$G = (V, E, c)$$

Here $V$ is the set of vertices or nodes of $G$; $E$ is the set of edges of $G$, each of which has the beginning and the end; $c: E \to V \times V$ is the edge-node *adjacency* or *incidence function*. This function assigns each edge to a pair of vertices so that the beginning of each edge is connected to the first element in the corresponding pair of vertices and the end of the same edge is connected to the second element in the same pair of vertices.

A multigraph is a graph when $c$ is an injection (Berge, 1973)**.**

Open systems demand a more general construction.

**Definition 2.4.** A *generalized oriented* or *directed multigraph G* has the following form:

$$G = (V, E, c: E \to (V \times V \cup V_b \cup V_e))$$

Here $V$ is the set of vertices or nodes of $G$; $E$ is the set of edges of $G$ (with fixed beginnings and ends); $V_b \approx V_e \approx V$; $c$ is the edge-node adjacency function, which assigns each edge either to a pair of vertices or to one vertex. In the latter case, when the image $c(e)$ of an edge $e$ belongs to $V_b$, it means that $e$ is connected to the vertex $c(e)$ by its beginning. When the image $c(e)$ of an edge $e$ belongs to $V_e$, it means that $e$ is connected to the vertex $c(e)$ by its end. Edges that are mapped to the set $V_b \cup V_e$ are called *open*.

The difference between multigraphs and generalized oriented multigraphs is that in a multigraph each edge connects two vertices, while in a generalized multigraph an edge may be connected only to one vertex.

A grid automaton is realized (situated) on grid. Here is an exact definition of this grid.

**Definition 2.5.** The *grid* G(*A*) of a grid automaton *A* is the generalized oriented multigraph that has exactly the same vertices and edges as *A*, while its adjacency function $c_{G(A)}$ is a composition of functions $p_{IA}$ and $c_A$, namely, $c_{G(A)}(l) = p_{IA}*(c_A(l))$ where $l$ is an arbitrary link from $C_A$, $A'_A$ and $A''_A$ are disjoint copies of $A_A$, and $p_{IA}* = (p_{IA} \times p_{IA}) * p_{IA} * p_{IA} : (\mathbf{P}_{IAin} \times \mathbf{P}_{IAout}) \cup \mathbf{P}_{IAin} \cup \mathbf{P}_{IAout} \to (A_A \times A_A) \cup A'_A \cup A''_A$.

Here $\times$ is the product and $*$ is the coproduct of mappings in the sense of category theory (Herrlich and Strecker, 1973).

**Example 2.4.** The grid G(**GA**) of the grid automaton **GA** from Example 2.1 is given in Figure 2.2.

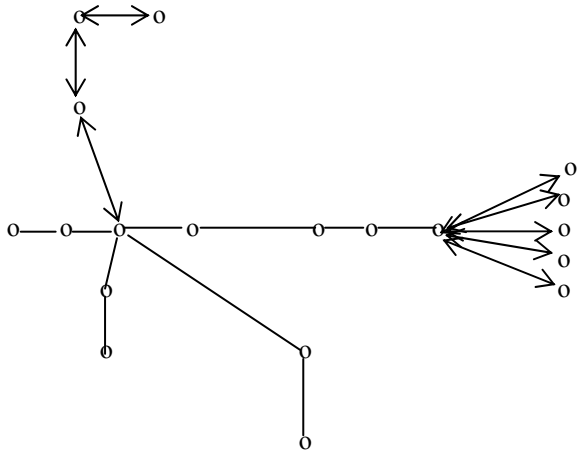

**Figure 2.2.** The grid of the grid automaton **GA** from Fig. 2.1.

Grids of the grid automata allow one to characterize definite classes of grid automata.

**Proposition 2.1.** A grid automaton $B$ is closed if and only if its grid G($B$) satisfies the condition Im $c \subseteq V \times V$, or in other words, the grid G($B$) of $B$ is a conventional multigraph.

Many classical models of computation, e.g., Turing machines, are closed grid automata.

**Proposition 2.2.** A grid automaton $B$ is an acceptor only if it has external input ports or/and Im $c \cap V_e \neq \emptyset$, i.e., the grid G($B$) has edges connected by their end.

**Proposition 2.3.** A grid automaton *B* is a transmitter only if it has external output ports or/and Im $c \cap V_b \neq \emptyset$, i.e., the grid G(*B*) has edges connected by their beginning.

**Proposition 2.4.** A grid automaton *B* is a transducer if and only if it has external input and output ports or/and Im $c \cap V_b \neq \emptyset$ and Im $c \cap V_e \neq \emptyset$, i.e., the grid G(*B*) has edges connected by their beginning and edges connected by their end.

**Definition 2.6.** The *connection grid* CG(*A*) of a grid automaton *A* is the generalized oriented multigraph nodes of which bijectively correspond to the internal ports of *A*, while edges and the adjacency function $c_{CG(A)}$ are the same as in *A*.

**Proposition 2.5.** The grid G(*B*) of a grid automaton *B* is a homomorphic image of its connection grid CG(*B*).

Indeed, by the definition of a grid automaton, ports are uniquely assigned to nodes (automata), and by the definition of the grid G(*B*) a grid automaton *B*, the adjacency function $c_{G(B)}$ of the grid G(*B*) is a composition of the port assignment function $p_B$ and the adjacency function $c_B$ of the automaton *B*.

**3. Elements of schema theory for interaction with the world**

Kant was perhaps the first to introduce (1781) the word schema into philosophy (Arbib, 1995; 2005; D'Andrade, 1995). For example, he describes the "dog" schema as a mental pattern that can delineate the figure of a four-footed animal in a general manner, without limitation to any single determinate figure from experience, or any possible image that a person can represent directly. The notion of a schema was introduced to neuroscience by Head and Holmes (1911) who discussed body schemas in the context of brain damage. Bartlett (1932) implemented the notion of a schema as part of a study of remembering. Another important use of schemas in psychology was initiated by Piaget, who viewed cognitive development from biological perspective and described it in terms of operation with schemas. With respect to adaptation, Piaget believed that humans desire a state of cognitive balance or equilibration. When the child experiences cognitive conflict (a discrepancy between what the child believes the state of the world to be and what she or he is experiencing) adaptation is achieved through assimilation and/or accommodation.

Assimilation involves making sense of the current situation in terms of previously existing structures or schema. Accommodation involves the formation of new mental structures or schema when new information does not fit into existing structures (e.g., a child encounters a skunk for the first time and learns that it is different from "dogs" and "cats." She must create a new schema for "skunks"). According to Piaget organization refers to the mind's natural tendency to organize information into related, interconnected structures. Piaget's notion of a "scheme" (the generalizable characteristics of an action that allow the application of the same action to a different context) is akin to Pierce's notion of a "habit" (a set of operational rules that, by exhibiting both stability and adaptability, lends itself to an evolutionary process). Both assume that schemas are adapted to yield successive levels of a cognitive hierarchy. Categories are not innate, they are constructed through the individual's experience. What is innate is the process that underlies the construction of categories.

The main source for this section is Arbib's (1985; 1989; 1992; 1995) version of schema theory, which the basic concept of which is *interaction schema* and which he has applied to the visuomotor coordination of the frog, high-level visual recognition, hand control and language processing – as well as perceptual robotics.

In this approach, interaction schemas are ultimately defined by the execution of tasks with the physical environment. A set of basic *motor schemas* is hypothesized to provide simple prototypical patterns of interaction with the world, whereas *perceptual schemas* recognize certain possibilities of interaction or other regularities of the physical world with various schema parameters representing properties such as size, location, and motion. Motor schemas are akin to control systems but distinguished in that they can be combined to form coordinated control programs that control the phasing in and out of patterns of co-activation, with mechanisms for the passing of control parameters from perceptual to motor schemas. These combine with *perceptual schemas* to form *assemblages* or *coordinated control programs* that interweave their activations in accordance with the current task and sensory environment to mediate more complex behaviors. A perceptual schema embodies the process that allows the organism to recognize a given domain of interaction. Various schema parameters represent properties such as size, location, and motion.

Many schemas may be abstracted from the perceptual-motor interface. Schema activations are largely task-driven, reflecting the goals of the organism and the physical and functional requirements of the task.

While work on schemas has to date yielded no efficient formalism, we do see the evolution of a theory of schemas as "programs" (in a generalized sense) for a system which has continuing perception of, and interaction with, its environment, with concurrent activity of many different schemas passing messages back and forth for the overall achievement of some goal. A schema is a self-contained computing agent (object) with the ability to communicate with other agents, and whose functionality is specified by some behavior. When we turn to brain theory, we further require that the schemas be implemented in specific neural networks.

A schema is both a store of knowledge and the description of a process for applying that knowledge. As such, a schema may be instantiated to form multiple schema *instances* as active copies of the process to apply that knowledge. E.g., given a schema that represents generic knowledge about some object, we may need several active instances of the schema, each suitably tuned, to subserve our perception of a different instance of the object. Schemas can become *instantiated* in response to certain patterns of input from sensory stimuli or other schema instances that are already active.

The alternative view (Arbib & Liaw, 1995) is that there is a limited set of schemas (maybe only one) and that only the schemas can be active. By contrast a schema instance is rather a record in working memory that records that a certain "region of space time" $\mathcal{R}$ activated a specific schema $S$ with certain parameters $\{P\}$ and confidence level $C$. On the latter view, processes of attention phase the activity of a schema in and out for different regions. Presumably, however, the working memory provides top-down activation of a schema when attention returns to those regions where the schema was recently active. These ideas should be tested by extending our formalism to address the various models – including the current outline of the latest – (cf. ( Itti & Arbib 2005)).

Each instance of a schema has an associated *activity level*. That of a perceptual schema represents a "confidence level" that the object represented by the schema is indeed present; while that of a motor schema may signal its "degree of readiness" to control some course of action. The activity level of a schema instance may be but one of many parameters that characterize it. Thus the perceptual schema for ''ball'' might include parameters to represent size, color, and velocity.

The use, representation, and recall of knowledge is mediated through the activity of a network of interacting computing agents, the schema instances, which between them provide processes for going from a particular situation and a particular structure of goals and tasks to a suitable course of action (which may be overt or covert, as when learning occurs without action or the animal changes its state of readiness). This activity may involve passing of messages, changes of state (including activity level), instantiation to add new schema instances to the network, and deinstantiation to remove instances. Moreover, such activity may involve self-modification and self-organization.

The key question is to understand how local schema interactions can integrate themselves to yield some overall result without explicit executive control, but rather through *cooperative computation,* a shorthand for ''computation based on the competition and cooperation of concurrently active agents''. For example, in interpretation of visual scenes, schema instances are used to represent hypotheses that particular objects occur at particular positions in a scene, so that instances may either represent conflicting hypotheses or offer mutual support. Cooperation yields a pattern of "strengthened alliances" between mutually consistent schema instances that allows them to achieve high activity levels to constitute the overall solution of a problem; competition ensures that instances which do not meet the evolving consensus lose activity, and thus are not part of this solution (though their continuing subthreshold activity may well affect later behavior). In this way, a schema network does not, in general, need a top-level executor, since schema instances can combine their effects by distributed processes of competition and cooperation, rather than the iteration of an inference engine on a passive store of knowledge. This may lead to apparently emergent behavior, due to the absence of global control.

In brain theory, a given schema, defined functionally, may be distributed across more than one brain region; conversely, a given brain region may be involved in many interaction schemas. A top-down analysis may advance specific hypotheses about the localization of (sub)-schemas in the brain and these may be tested by lesion experiments, with possible modification of the model (e.g., replacing one schema by several interacting schemas with different localizations) and further testing.

Schemas, and their connections within a schema network, must change so that over time they may well be able to handle a certain range of situations in a more adaptive way. In a general setting, there is no fixed repertoire of basic schemas. New schemas may be formed as assemblages of old schemas; but once formed a schema may be tuned by some adaptive mechanism. This tunability of schema assemblages allows them to become "primitive", much as a skill is honed into a unified whole from constituent pieces. Such tuning may be expressed at the level of schema theory itself, or may be driven by the dynamics of modification of unit interactions in some specific implementation of the schemas. The theory of interaction schemas is consistent with a model of the brain as an evolving self-configuring system of interconnected units.

Once an interaction schema–theoretic model of some animal behavior has been refined to the point of hypotheses about the localization of schemas, we may then model a brain region by seeing if its known neural circuitry can indeed be shown to implement the posited schema. In some cases, the model will involve properties of the circuitry that have not yet been tested, thus laying the ground for new experiments. In AI, individual schemas may be implemented by artificial neural networks, or in some programming language on a "standard" (possibly distributed) computer.

Schema theory is far removed from the serial symbol-based computation. Increasingly, work in Al now contributes to schema theory, even when it does not use this term. For example, Minsky (1986) espoused a *Society of Mind* analogy in which "members of society", the agents, are analogous to schemas. The study of interactive "agents" more generally has become an established theme in AI. Their work shares with schema theory, with its mediation of action through a network of schemas, the point that no single, central, logical representation of the world need link perception and action - the representation of the world is *the pattern of relationships* bet*ween all its partial representations*. Another common theme is the study of the "evolution" of simple "creatures" within increasingly sophisticated sensorimotor capacities.

Here are some examples of interaction schemas.

**Example 3.1.** The schema of face recognition is tentatively acquired at around two or three months of age (which succeeds to a previous scheme already present at birth) This schema corresponds to the mental structure which connects the various states of a face defined by configurations of perceptual indices (front view, side view, etc.) related to actions-transformations (head rotations, subject's or object's rotation).

**Example 3.2.** The schema of (shape or) size constancy is the insertion of the various sizes of an object related to its distance from the perceiver in a transformational system (system of transformations) governing the moves of the object. Present at birth, it could be reconstructed during the first months of life.

**Example 3.3.** The schema of object's permanence (the "objective" form), the one achieved according to Piaget at around 16 to 18 months of age, is the mental structure which connects the various successive states of a set of objects (their different localizations or relative positions) to their successive displacements (transformations), even bridging across periods when the object disappears from the view.

To achieve better comprehensibility, interaction schemas are usually represented in a graphical form as in Figure 3.1.

**Example 3.4.** A hypothetical coordinated control schema for reaching and grasping is given in Figure 3.1 (Arbib, 1989).

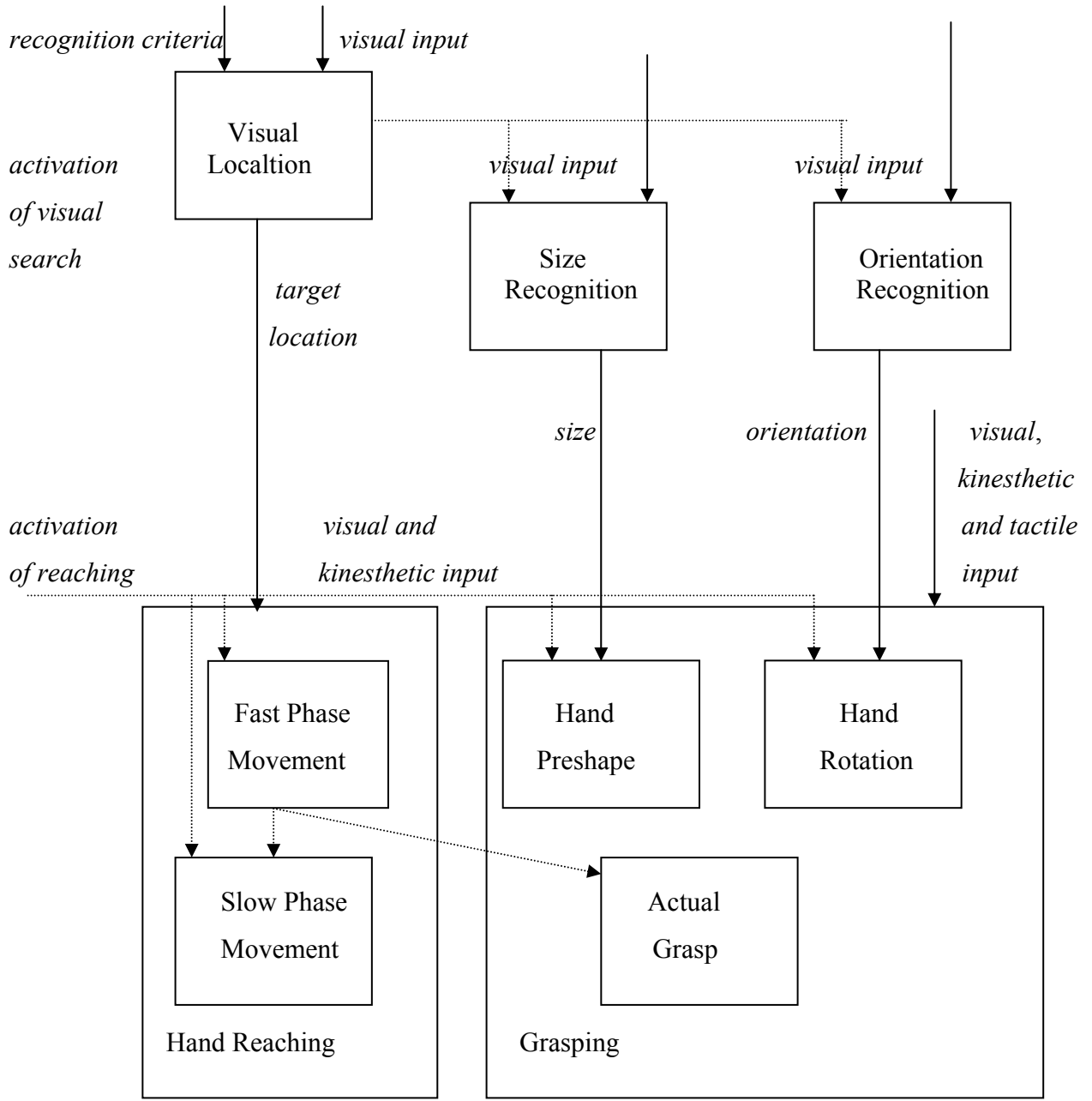

**Figure 3.1.** Dashed lines – activation of signals (i.e., control links, in our terminology); solid lines – transfer of data (i.e., information links, in our terminology).

Interaction schemas form the base for the Computational Neuroscience, structure and function of which are given in Fig. 3.2. It is interesting to note that even subneural modeling brings us to grid automata.

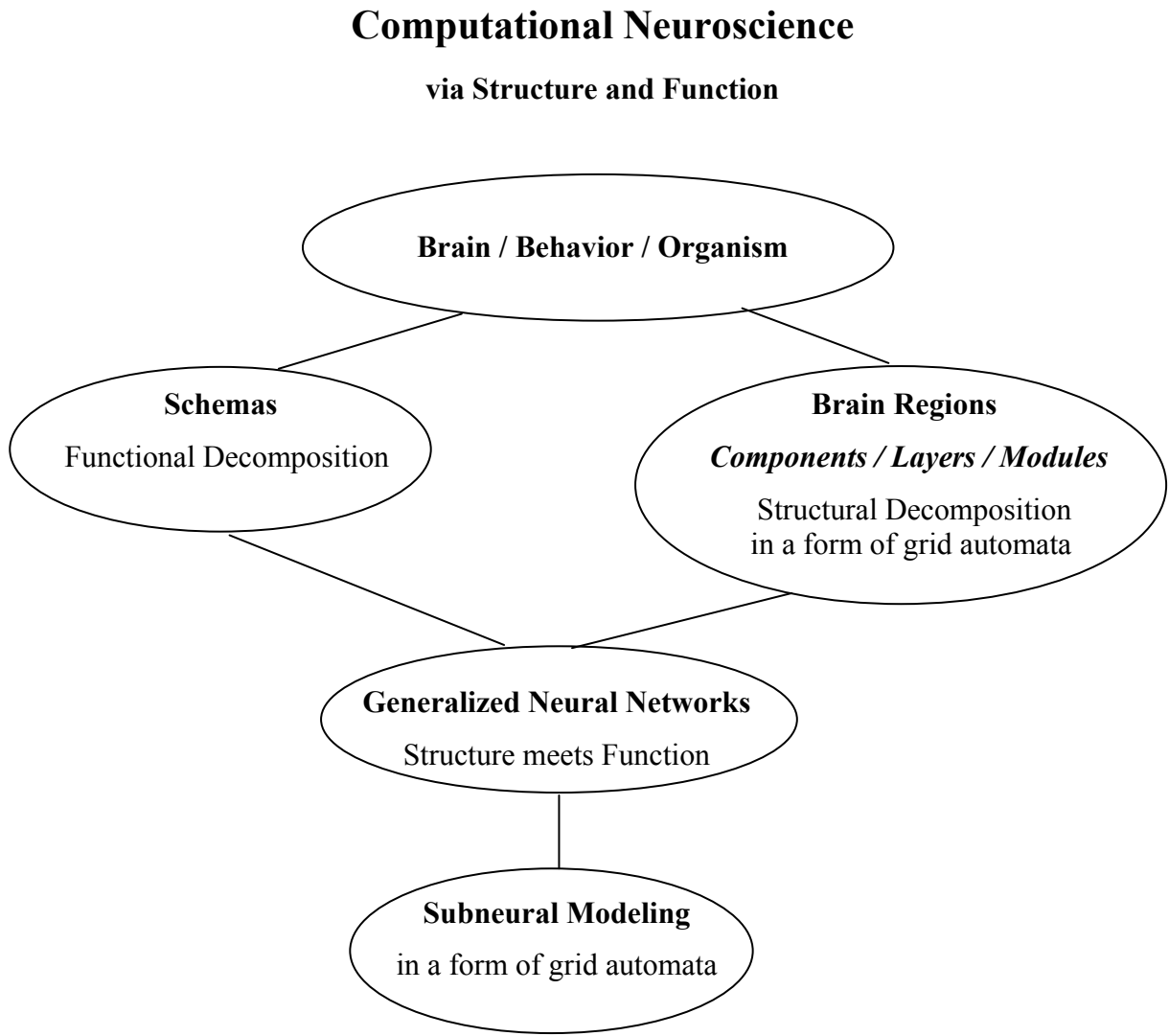

**Figure 3.2.** A version of the schema for the Computational Neuroscience suggested by M. A. Arbib

## 4. Other notions of schema

A notion of a schema rather different in emphasis and properties from those we have just been considering has been very popular in programming, where it was formalized and extensively used for theoretical purposes. At the beginning, program schemas, or program schemata, were introduced by A.A. Lyapunov in 1953 and published later in (Lyapunov, 1958) under the name "operator schema." Afterwards Ianov, a graduate student of Lyapunov, transformed operator schemas into a logical form called a *logical schema of algorithm* (later named *Ianov program schemata*) and proved many properties of these schemas (Ianov, 1958; 1958a; 1958b). The main result of Ianov is a theorem about the decidability of equivalency of schemas that use only one-argument functions. Approximately at the same time, Kaluznin (1959) introduced the concept of a graph-schema of an algorithm. Subsequently, this concept was generalized by Bloch (1975) and applied to automaton synthesis, discrete system design, programming, and medical diagnostics.

Program schemas were later studied by different authors, who introduced various kinds of program schemas: recursive, push-down, free, standard, total schemas (cf., for example, (Karp and Miller, 1969; Paterson, and Hewitt, 1970; Garland and Luckham, 1971; Logrippo, 1978)). Fischer (1993) introduced the mathematical concept of a lambda-calculus schema to compare the expressive power of programming languages. The theory of program schemas has been considered as a base for (Yershov, 1977) or one of the main directions (Kotov, 1978) in theoretical programming. In the 1960s, program schemas were used to create programming languages and build translators. To study parallel computations, flow graph and dataflow schemas have been introduced and utilized (Slutz, 1968; Keller, 1973; Dennis, Fossen, and Linderman, 1974). Dataflow schemas are formalizations of dataflow languages. Program schemas and dataflow schemas formed an implicit base for the development of the first programming metalanguage (Burgin, 1973; 1976).

Moreover, the advent of the Internet and introduction of the Extensible Markup Language, abbreviated XML, started the development of schema languages (cf., for example, (Duckett, *et al*, 2001; Van Der Vlist, 2004)). As developers know, the advantage of XML is that it is extensible, even to the point that you can invent new elements and attributes as you write XML documents. Then, however, you need to define your changes so that applications will be able to make sense of them and this is where XML schema languages come into play. In these languages, schemas are machine-processable specifications that define the structure and syntax of metadata specifications in a formal schema language. There are many different XML schema languages (W3C Schema, Schematron, Relax NG, and so on). They are based on schemas that define the allowable content of a class of XML documents. Schema languages form an alternative to the DTD (Document Type Definition), and offer more powerful features including the ability to define data types and structures. XML schemas from these languages provide means for defining the structure, content and semantics of XML documents, including metadata. A specification for XML schemas is developed and maintained under the auspices of the World Wide Web Consortium. The Resource Description Framework (RDF) is an evolving metadata framework that offers a degree of semantic interoperability among applications that exchange machine-understandable metadata on the Web. RDF Schema is a specification developed and maintained under the auspices of the World Wide Web Consortium. The *Schematron* schema language differs from most other XML schema languages in that it is a rule-based language that uses path expressions instead of grammars. This means that instead of creating a grammar for an XML document, a Schematron schema makes assertions applied to a specific context within the document. If the assertion fails, a diagnostic message that is supplied by the author of the schema can be displayed. RELAX NG is a grammar-based schema language, which is both easy to learn for schema creators and easy to implement for software developers.

A natural tool for providing flexible data structures is to create schemas. Such schemas are used to describe an object and any of the interrelationships that exist within a data structure. There are many different kinds of schema used in different areas of information technology. For example, relational databases such as SQL Server use schemas to contain their table names, column keys, and provide a repository for trigger and stored procedures. Also when a developer creates a class definition, he or she can define schemas to provide the object-oriented interface to properties, methods, and events.

An XML schema is by definition a well-formed XML document. At the top of an XSD file is a set of namespaces. These are an optional set of declarations that provide a unique set of identifiers that associate a set of XML elements and attributes together. The original namespace in the XML specification was released by the W3C as a URI-based way to differentiate various XML vocabularies. This was then extended under the XML schema specification to include schema components and not just single elements and attributes. The unique identifier was redefined as a URI that doesn't point to a physical location, but to a security boundary that is owned by the schema author. The namespace is defined through two declarations - the XML schema namespace and target namespace.

Special kind of XML schemas has been developed for energy simulation data representation (Gowri, 2001). Another application of XML schemas is e-business. For instance, the ebXML specification schema developed by UN/CEFACT and Oasis provides a standard framework by which business systems may be configured to support execution of business collaborations, which consist of business transactions (cf. Business Process Specification Schema). Transactions can be implemented using one of many available standard patterns. These patterns determine the actual exchange of business documents and signals between the partners to achieve the required electronic commerce transactions.

Another example is the XML schema definition developed by the Danish Broadcasting Corporation for business-to-business exchange interface defined the DR metadata standard.

Star Schema determines a method of organizing information in a data warehouse that allows the business information to be viewed from many perspectives. XML schemas are used for modeling business objects (Daum, 2003).

In addition, an important tool in database theory and technology is the notion of the database schema, which gives a general description of a database, is specified during database design, and is not expected to change frequently (Elmasri and Navathne, 2000). Database schemas are represented by schema diagrams. Database management system (DBMS) architecture is often specified utilizing database schemas. Three important tasks of databases are (Elmasri and Navathne, 2000):

1. Insulation of program and data (program-data and program-operation independence).
2. Support of multiple user views.
3. Use of a catalog to store the database description (schema).

To realize these tasks, the three-schema architecture, or ANSI/SPARC architecture, of

DBMS was developed (Tsichridsis and Klug, 1978). In this architecture, schemas are defined at three levels:

1. The *internal level* has an *internal schema*, which describes the physical storage structure of the database.
2. The *conceptual level* has a *conceptual schema*, which describes the structure of the whole database for a community of users.
3. The *external* or *view level* includes a number of *external schemas* or *user views*. Each external schema describes the part of the database that a particular user group is interested in.

Most DBMS do not separate the three levels completely, but support the three-schema architecture to some extent.

The mathematical schema theory developed in this paper encompasses all types of schemas used in programming, database theory, and computer science.

A notion of a schema has been also used in mathematical logic, metamathematics, and set theory. Von Neumann (192**7**) introduced the concept of an axiom schema. It has become very useful in axiomatic set theories (for instance, the axiom of subsets is, according to conceptions of Skolem, Ackermann, Quine and some other logicians, an axiom schema) and other axiomatic mathematical theories (Fraenkel and Bar-Hillel, 1958). Axiomatizability by a schema was studied in the context of general formal theories (Vaught, 1967). In addition to axiom schemas, schemas of inference (e.g., syllogism schemas) have been also studied in mathematical logic (cf., for example, (Fraenkel and Bar-Hillel, 1958)). Actually, syllogisms introduced by Aristotle, as well as deduction rules of modern logic are schemas for logical inference and mathematical proofs.

Another mathematical field where the concept of schema is used is category theory. This concept was introduced by Grothendieck (1957) in a form equivalent to a multigraph and later generalized to the form of a small category. From categories, the concept of a schema came to algebraic geometry, where now it play an essential role.

The mathematical schema theory developed in this paper encompasses all types of schemas used in mathematical logic, metamathematics, and set theory. However, it is necessary to stress that here interaction schemas are our main concern.

## 5. Elements of a mathematical schema theory

The first step to formalization of interaction schemas was made by creation of the RS (Robot Schema) language (Lyons, 1986; Lyons and Arbib, 1989) and NSL (Neural Simulation Language) (Weitzenfeld, 1989; Weitzenfeld, Arbib, and Alexander, 2002). RS is a language designed to facilitate sensory-based task-level robot programming. RS uses port automata (Arbib, Steenstrup and Manes 1983) to provide semantics of schemas. NSL was developed to aid the exploration of neural network simulations through interactive computer graphics. Arbib and Ehrig (1990) made two first attempts at providing a rapprochement between a methodology for parallel and distributed computation in the context of brain theory and perceptual robotics based on RS-schemas and an algebraic category theory of the specification of modules and their interconnections developed in (Blum, Ehrig, and Parisi-Presicce, 1987; Ehrig, and Mahr, 1985; 1990). However, as Arbib (2005) writes: "It must be confessed that [that work] was more a program for research than a presentation of results, and that research remains to be done."

Here we continue this research and develop a mathematical schema theory that enables us to represent not only external features of interaction schemas and their functioning but also essential structural peculiarities of interaction schemas and their assemblages. At first, we develop a general mathematical concept of schema and later it is specialized so as to achieve a mathematical model for interaction schemas. To reach sufficient generality, we build a general concept of schema by transformation of the grid automaton structure changing some of the automaton elements to variables.

**Remark 5.1.** This understanding encompasses both interaction and formal schemas. In formal schemas, variables are represented by their names in a conventional manner (cf. Examples 5.1 – 5.9). In informal schemas, variables are represented by their descriptions or specifications in the form of a text (cf. Example 3.4), picture, text with pictures, etc.

The transition from the theory of grid automata to the mathematical schema theory is comparable to the transition from numbers to functions.

Some properties of schemas are similar to properties of grid automata, while others are essentially different. It is possible to consider grid automata as schemas of the zero level. In practice, grid automata are realizations of schemas.

Similar to grid automata, schemas also can have ports, which are specific schema elements which belong (are assigned) to schema nodes and through which information/data come into (*output ports* or *outlets*) and are sent outside the schema (*input ports* or *inlets*). Thus as before, any system ***P*** of ports is the union of its two disjunctive subsets ***P*** = ***P***$_{in}$ ∪ ***P***$_{out}$ where ***P***$_{in}$ consists of all inlets from ***P*** and ***P***$_{out}$ consists of all outlets from ***P***. If there are ports that are both inlets and outlets, we combine such ports from couples of an input port and an output port.

To formalize schemas, we consider, at first, those elements from which schemas are built of. There are three types of schema elements: *nodes* or *vertices*, *ports*, and *ties* or *edges*. Elements of all types belong to three classes:

1. Automaton/node, port, and connection/edge *constants*.
2. Automaton/node, port, and connection/edge *variables*.
3. Automata, ports, and connections *with variables*.

**Example 5.1.** The symbol *T* can be used as an automaton variable the range of which is the class of Turing machines. The expression *NN* can be used as an automaton variable the range of which is the class of neural networks. The symbol *P* can be used as an automaton variable the range of which is the class of port automata. Thus, variables T for Turing machines, A for finite automata, N for neural networks, etc. in the schema from Example 5.2 are automaton/node variables.

**Example 5.2.** Information connections denoted by solid lines and process connections denoted by dashed lines in the schema from Example 5.3 are connection/edge variables.

It is also possible to use different connection variables for links implemented on physical media, such as coaxial cable, twisted pair, or optical fiber.

**Example 5.3.** The expression *T*[with *x* tapes] can be used as a denotation for an automaton with the variable *x* the range of which is the number of Turing machines tapes.

**Example 5.4.** The expression *c*[with bandwidth *x*] can be used as a denotation for a connection/link with the variable *x* the range of which is the bandwidth (throughput) of the link.

**Remark 5.2.** Each variable *x* is determined by its name *x* and range Rg *x*. Types of ranges determine types of variables. For instance, a variable whose range encompasses some class of neural networks has the neural network type.

**Remark 5.3.** Variables in a schema form in a general case not a set but a multiset (cf., for example, Knuth, 1997; 1998) because the same variable $x$ may be assigned to different nodes, links or ports.

In addition to variables, we need variable functions. A *variable function* takes values in variables. For instance, a linear real function $f$ is a variable function as it has the form $f(x) = ax + b$ where $a$ and $b$ are arbitrary real numbers. Another example of a variable function is the function that takes any value $x^n$ for a given argument $x$.

Variable functions can be of different types:

*fuzzy functions* in the sense of fuzzy set theory when values of the function have estimates, e.g., to what extent this value is correct, true or exact (Zimmermann, 1991);

*nondeterministic functions* when values of the function are not uniquely determined by the argument;

*probabilistic functions* in the sense of fuzzy set theory when values of the function have probabilities showing, e.g., to what extent this value is correct, true or exact.

Wave function in quantum mechanics is an example of a probabilistic function.

**Remark 5.4.** There is one-to-one correspondence between nondeterministic functions and set-valued functions, in which values are some sets.

**Remark 5.5.** There are fuzzy functions with fuzzy domain and/or range. However, here we do not consider such functions.

**Remark 5.6.** There is one-to-one correspondence between fuzzy functions in the above sense and fuzzy-set-valued functions, in which values are some fuzzy sets.

All these structures make possible to define basic schemas.

**Definition 5.1.** A *basic schema R* is the following system that consists of two sets, two multisets, and one mapping

$R = (A_R, \mathbf{V}_{NR}, C_R, \mathbf{V}_{CR}, c_R)$

Here:

The set $A_R$ is the set of all automata from $R$;

the multiset $\mathbf{V}_{NR}$ consists of all automaton variables from $R$;

the set $C_R$ is the set of all connections/links from $R$;

the multiset $\mathbf{V}_{CR}$ consists of all link variables from $R$;

and

$c_R : C_R \cup V_{CR} \to ((A_R \cup V_{NR}) \times (A_R \cup V_{NR})) \cup (A'_R \cup V'_{NR}) \cup (A''_R \cup V''_{NR})$ is a (variable) function, called the *node-link adjacency function*, that assigns connections to nodes where $A'_R$ and $A''_R$ are disjunctive copies of $A_R$, while $V'_{NR}$ and $V''_{NR}$ are disjunctive copies of $V_{NR}$.

In some cases, we need more information about schemas. A specific kind of such information is related to ports of the nodes. Ports are used to provide necessary connections between nodes inside the schema and between the schema and other systems. In this case, we consider port schemas.

**Definition 5.2.** A *port schema B* is the following system that consists of three sets, three multisets, and three mappings

$B = (A_B, V_{NB}, P_B, V_{PB}, C_B, V_{CB}, p_{IB}, c_B, p_{EB})$

Here:

The set $A_B$ is the set of all automata from $B$;

the multiset $V_{NB}$ consists of all automaton variables from $B$;

the set $C_B$ is the set of all connections/links from $B$;

the multiset $V_{CB}$ consists of all link variables from $B$;

the set $P_B = P_{IB} \cup P_{EB}$ (with $P_{IB} \cap P_{EB} = \varnothing$) is the set of all ports of $B$, $P_{IB}$ is the set of all ports (called *internal ports*) of the automata from $A_B$, and $P_{EB}$ is the set of *external ports* of $B$, which are used for interaction of $B$ with different external systems;

the multiset $V_{PB}$ consists of all port variables from $B$ and is divided into two disjunctive submultisets $V_{PBin}$ that consists of all variable inlets from $B$ and $V_{PBout}$ consists of all outlets from $B$;

$p_{IB} : P_{IB} \cup V_{PB} \to A_B \cup V_{NB}$ is a (variable) total function, called the *internal port assignment function*, that assigns ports to automata;

$c_B : C_B \cup V_{CB} \to ((P_{Ibout} \cup V_{PBout}) \times (P_{Ibin} \cup V_{PBin})) \cup (P'_{IBin} \cup V'_{PBin}) \cup (P'_{IBout} \cup V'_{PBout})$ is a (variable) function, called the *port-link adjacency function*, that assigns connections to ports where $P'_{IGin}$, $P''_{Igout}$, $V'_{PBin}$ and $V'_{PBout}$ are disjunctive copies of $P'_{IGin}$, $P''_{Igout}$, $V'_{PBin}$ and $V'_{PBout}$, correspondingly;

and

$p_{EB} : \boldsymbol{P}_{EB} \cup \mathbf{V}_{PB} \to \boldsymbol{A}_B \cup \boldsymbol{P}_{IB} \cup \boldsymbol{C}_B \cup \mathbf{V}_{NB} \cup \mathbf{V}_{PB} \cup \mathbf{V}_{CB}$ is a (variable) function, called the *external port assignment function*, that assigns ports to different elements from *B*.

Usually, basic schemas are used when the modeling scale is big, i.e., at the coarse-grain level, while port schemas are used when the modeling scale is small and we need a fine-grain model.

Schemas without ports, i.e., basic schemas, give us the first approximation to cognitive structures, while schemas with ports, i.e., port schemas, is the second (more exact) approximation. In some cases, it is sufficient to use schemas without ports, while in other situations to build an adequate, flexible and efficient model, we need schemas with ports. For instance, interaction schemas (Arbib, 1985), schemas of programs (cf. Garland and Luckham, 1969; Dennis, et al, 1974; Fischer, 1993), or flow-charts (cf. Burgin, 1976; 1985; 1996) do not traditionally have ports. Even schemas of computer hardware are usually presented without ports (Heuring and Jordan, 1997).

**Definition 5.3.** Internal ports of a port schema *B* to which no links are attached are called *open*. External ports of a port schema *B* to which no links or automata are attached are called *free*.

External ports of a port schema *B*, being always open, are used for connecting *B* to some external systems.

**Remark 5.7.** It is possible to consider two representations of schemas: planar or graphical and linear or symbolic. To achieve better comprehension, schemas are usually represented in a graphical form as in Figures 5.1 – 5.5.

**Example 5.5.** A basic schema, concretization of which is the grid automaton from Figure 2.1, is given in Figure 5.1.

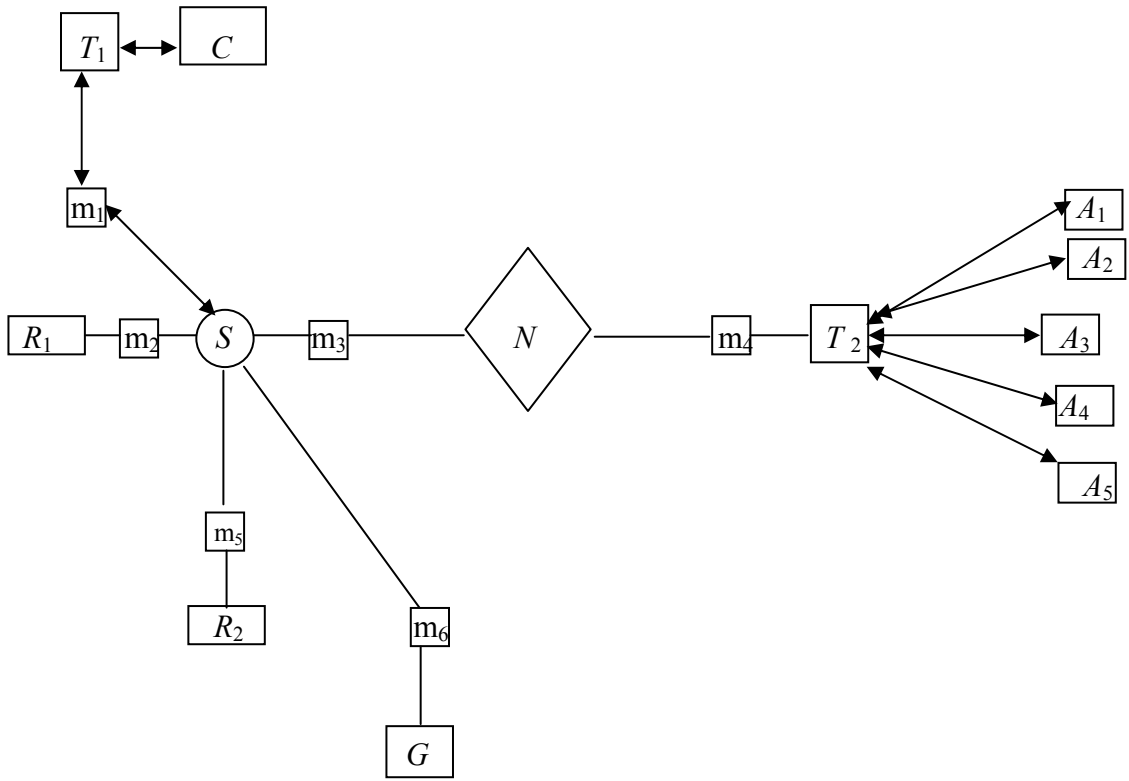

**Figure 5.1.** A schema of the grid automaton **GA** from Fig. 2.1.
  $T_i$ is a variable the range of which is the class of all Turing machines;
  $R_i$ is a variable the range of which is the class of all random access machines;
  $S$ is a variable the range of which is the class of all servers;
  $m_i$ is a variable the range of which is the class of all modems;
  $N$ is a variable the range of which is the class of all neural networks;
  $A_i$ is a variable the range of which is the class of all finite automata;
  $C$ is a variable the range of which is the class of all cellular automata;
  $G$ is a variable the range of which is the class of all grid automata.

In the schema from Figure 5.1, variables form the multiset that contains: two variables $T$, one variable $N$, two variables $R$, five variables $A$, one variable $C$, six variables $m$, one variable $C$, and one variable $G$.

**Example 5.6.** A formal basic schema that formalizes the interaction schema from Figure 3.1 is given in Figure 5.2. This schema 5.2 has connections/links of two types: links for activation of nodes and for transfer of data. Such a formalization of the schema from Figure 3.1 allows us to better study its properties and transformations. It demonstrates that this schema has realizations not only by the brain neural structures but also by computer programs.

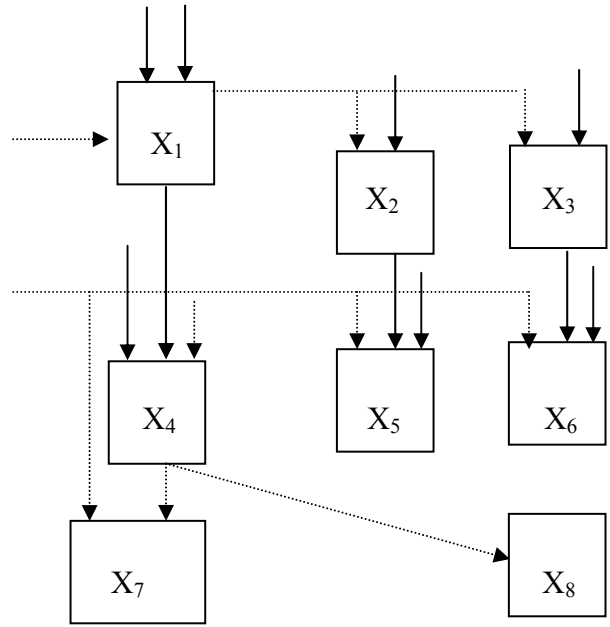

**Figure 5.2.** Dashed lines – activation of signals; solid lines – transfer of data. $X_1$ is a variable the range of which is the class of all schemas (algorithms or neural assemblages) for visual location; $X_2$ is a variable the range of which is the class of all schemas (algorithms or neural assemblages) for size recognition; $X_3$ is a variable the range of which is the class of all schemas (algorithms or neural assemblages) for orientation recognition; $X_4$ is a variable the range of which is the class of all schemas (algorithms or neural assemblages) for fast phase movement; $X_5$ is a variable the range of which is the class of all schemas (algorithms or neural assemblages) for hand preshape; $X_6$ is a variable the range of which is the class of all schemas (algorithms or neural assemblages) for hand rotation; $X_7$ is a variable the range of which is the class of all schemas (algorithms or neural assemblages) for slow phase movement; $X_8$ is a variable the range of which is the class of all schemas (algorithms or neural assemblages) for actual grasp.

The following algorithm shows how to get basic schemas from port schemas:

The internal port assignment function and port-link adjacency function determine the *node-link adjacency function $nc_B$* of the port schema $B$ in the following way. Let $l \in C_B$, $\overline{P}_{IBin} = P_{IBin} \cup V_{PB}$, $\overline{P}_{IBout} = P_{IBout} \cup V_{PB}$, $\overline{A}_B = A_B \cup V_{NB}$, $\overline{A}'_B$ and $\overline{A}''_B$ are disjoint copies of $\overline{A}_B$, and $p_{IB}* = (p_{IB} \times p_{IB}) * p_{IB} * p_{IB} : (\overline{P}_{IBin} \times \overline{P}_{IBout}) \cup \overline{P}_{IBin} \cup \overline{P}_{IBout} \to (\overline{A}_B \times \overline{A}_B) \cup \overline{A}'_B \cup \overline{A}''_B$. Here $\times$ is the product and $*$ is the coproduct of mappings in the sense of category theory (cf., for example, Herrlich and Strecker, 1973). Then $nc_B$ is a composition of functions $p_{IB}$ and $c_B$, namely, $nc_B(l) = p_{IB}*(c_B(l))$.

The node-link adjacency function $nc_B$ determines a schema in which links are adjusted directly to nodes, ignoring ports. Thus, it is possible to exclude ports from the schema, obtaining a schema without ports or basic schemas. This algorithm gives us a basic schema, which is denoted by D$B$ where D$B = (A_B, V_{NB}, C_B, V_{CB}, nc_B)$

At the same time, it is possible to consider any basic schema as a special kind of port schemas, in which any node has exactly one port and all connections go through this port.

**Example 5.7.** A basic schema (from Burgin, 2005; Ch. 2) of a Turing machine $A_{T,w}$ that reduces the problem of deciding whether a given Turing machine has some non-trivial property $P$ to the halting problem for Turing machines. This example shows that it is possible to build schemas of algorithms and automata.

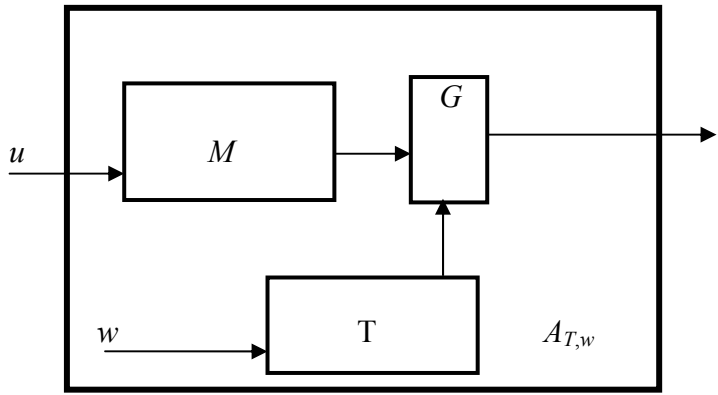

**Figure 5.3.** $T$ denotes some Turing machine from a given class **K**. $M$ denotes a Turing machine that does not have the property $P$. $G$ denotes a finite automaton with two inputs. One input come from the outside, while the second is the output of $T$. The automaton $G$ can be in two states: closed and open. Initially $G$ is closed until it receives some input from $T$, which makes it open. When $G$ is closed, it gives no output. When $G$ is open, it gives the word that comes to $G$ from $M$ as its output. The structure of the Turing machine $A_{T,w}$.

When an informal schema, such as an interaction schema or flow-chart of a program, is formalized, its formal representation is a mathematical model of this schema. This model allows one to study, build, and apply schemas utilizing powerful tools of mathematics. The procedure of formalization is rather simple. To get a formal representation of an interaction schema, we denote descriptions by variables, properly assign ranges of these variables, and make relevant substitutions in the schema.

**Remark 5.8.** It is possible to consider schemas with zero variables. Then any grid automaton, and consequently, any algorithm, becomes a schema.

**Remark 5.9.** If an automaton (system) is given by its specification in the sense of Blum, Ehrig, and Parisi-Presicce (1987) and Ehrig and Mahr (1985; 1990), then components and their compositions become a special kind of schemas defined in this paper. This allows one to more rigorously develop a component-based technology similar to one developed by these same authors.

Similar to a grid automaton, a port schema $P$ is described by three grid characteristics, three node characteristics, and three edge characteristics.

<u>The grid characteristics are:</u>

1. The *space organization* or *structure* of the schema $P$. This space structure may be in physical space, reflecting where the corresponding information processing systems (nodes) are situated, or it may be a mathematical structure defined by the geometry of node relations. Besides, we consider three levels of space structures: local, region, and global space structures of a schema. Sometimes these structures are the same, while in other cases they are different. The space structure of a schema can be *static* or *dynamic*. The dynamic space structure can be of two kinds: *persistent* or *flexible*. However, in contrast to grid automata, the space structure of a schema may be variable.

    Inherent structures of the schema are represented by its grid and connection grid (cf. Definitions 5.8 and 5.9). Due to a possible nondeterminism in the port assignment functions and port-link adjacency function, there is a possibility of nondeterminism in inherent structures of the schema .

2. The *topology* of the schema $P$ is a complex structure that consists of node topology determined by the type of the node neighborhood and port topology determined by the type of the port neighborhood. A neighborhood of a node (port) is the set of those nodes (ports) with which this node directly interacts (is directly connected). As the port assignment functions and port-link adjacency function may be nondeterministic, the topology of the schema *P* also may be nondeterministic. In particular, a schema may have fuzzy or probabilistic topology.

For deterministic schemas, we have three main types of topology:

- A uniform topology, in which neighborhoods of all nodes of the schema have the structure.

- A regular topology, in which the structure of different node neighborhoods is subjected to some regularity.

- An irregular topology where there is no regularity in the structure of different node neighborhoods.

An example of a regular but nonuniform schema topology is the schema of a cellular automaton in the hyperbolic plane or on a Fibonacci tree (Margenstern, 2002). In this schema, nodes are variables ranging over finite automata, while all edges/links are fixed.

Nondeterministic schemas can also be regular and irregular.

3. The *dynamics* of the schema *P* determines by what rules its nodes exchange information with each other and with a tentative environment of *P* and in particular, how nodes use ports and corresponding links. This dynamic is usually an algorithmic function that depends on values of its variables because some of nodes and/or links are variables and there is a permissible nondeterminism in the port assignment functions and port-link adjacency function.

Interaction with the environment separates two classes of schema: *open schemas* allow interaction (accepting and transmitting information) with the environment through definite connections, while *closed schemas* do not have means for such interaction. For instance, traditional schemas representing concepts and logical propositions are closed.

Existence of free ports makes a closed schema *potentially open* as it is possible to attach connections to these ports.

The node characteristics are:

1. The *type* and *structure* of the node, including structures of its ports. There are different levels of node typology. On the highest level, there are two types of nodes: an automaton node and a variable node. Each of these types has subtypes, e.g., a neural network, Turing machine or finite state machine. These subtypes form the next level of the type hierarchy. Subtypes of these subtypes (e.g., a Turing machine with one linear tape) form one more level of the type hierarchy and so on.

2. The *external dynamics* of the node determines interactions of this node. According to this characteristic, there are three types of nodes: *accepting nodes* that only accept or reject their input; *generating nodes* that only produce some input; and *transducing nodes* that both accept some input and produce some input. Note that nodes with the same external dynamics can have different dynamics when they work in a grid. For instance, let us take two nodes: a transducing node *B* and a generating node *B*. Initially they have different dynamics. However, as parts of a schema *P*, they both work as generating nodes because the schema dynamics prescribes this. For nodes of the schema that are variables, we have not a definite dynamics but a type of dynamics.

    Primitive ports do not change node dynamics. However, compound ports are able to influence processes in the whole schema and in the node to which they belong. For instance, a compound port can be an automaton or even a schema.

3. The *internal dynamics* of the node determines what processes go inside this node. For nodes of the schema that are variables, we have not a definite dynamics but a type of dynamics. For instance, it may be given that the node with number 3 in a schema computes function *f*(*x*). Such nodes are usually used in program schemas (which are traditionally called program schemata (cf., for example, (Fischer, 1993))).

The edge characteristics are:

1. The *external structure* of the edge. According to this characteristic, there are three types of edges: a *closed edge* (a link or link variable) both sides of which are connected to ports of the schema; an *ingoing edge* in which only the end side is connected to a port of the port schema; and an *outgoing edge* in which only the beginning side is connected to a port of the port schema

2. Properties and the *internal structure* of the edge. There are different levels of edge typology. On the highest level, there are two types: constant and variable links. Each of these types has subtypes that form the next level of the type hierarchy. According to the internal structure, there are three subtypes of edges: a *simple channel* that only transmits data/information; a *channel with filtering* that separates a signal from noise; and a *channel with data correction*. Subtypes of these subtypes form the next level and so on.
3. The *dynamics* of the edge determines edge functioning. For instance, two important dynamic characteristics of an edge are bandwidth as the number of bits (data units) per second transmitted on the edge and throughput as the measured performance of the edge. In schemas, these characteristics may be variable.

Properties of links/edges separate all links into three standard classes:

1. *Information link/connection* is a channel for processed data transmission.

2. *Control link/connection* is a channel for instruction transmission.

3. *Process link/connection* realizes control transfer and determines how the process goes by initiation of a node in the grid by another node (other nodes) in the grid.

Process links determine what to do, control links are used to instruct how to work, and information links supply automata with data in a process of schema or its instantiation functioning.

There are different types and kinds of schema variables.

The *dynamic typology* discerns three types of basic variables:

1. *System variables*.
2. *Function variables*.
3. *Process variables*.

The schema from Example 5.5 uses system variables ($T_i$ for Turing machines, $A_i$ for finite automata, $N$ for neural networks, etc.).

The schema from Example 5.6 uses function variables, e.g., $X_2$ is a variable for such function as size recognition.

The *scaling classification* discerns three types of variables:

1. *Individual variables* that are used in one node, port or link from the schema.

2. *Local variables* that are used in a group of nodes, ports or links from the schema.
3. *Global variables* that are used for the whole schema.

Difference between constants and variables in schemas results in existence of special classes of schemas:

- Basic/port schemas with constant nodes;
- Basic/port schemas with constant links;
- Port schemas with constant ports;
- Port schemas with constant port assignment;
- Basic/port schemas with constant node-link adjacency.
- Port schemas with dynamic port assignment;
- Basic/port schemas with dynamic node-link adjacency.
- Port schemas with deterministic port assignment;
- Basic/port schemas with deterministic node-link adjacency.

Let us consider operations on schemas. Utilization of different schemas usually involves various operations. There are three basic vertical unary operations in the hierarchy of both basic and port schemas: abstraction, concretization, and determination.

**Definition 5.4.** Changing a variable to a constant from the range of this variable is called an *interpretation* of this variable.

**Definition 5.5.** An operation of changing (interpreting) some of the variables in a schema *R* to constants is called a *concretization operation* Con applied to *R*, while the result Con *R* of this operation is called a *concretization* of *R*.

**Example 5.8.** An *instantiation* of a schema in the sense of (Arbib, 1989) is its maximal concretization.

**Example 5.9.** The grid automaton from Figure 2.1 is a concretization of the schema from Figure 5.1.

**Lemma 5.1.** Concretization of a schema preserves the schema topology and structure.

**Definition 5.6.** If a *concretization* Con *R* of a schema *R* is a grid automaton, then Con *R* is called a *realization* of *R*, while the corresponding operation is called a *realization operation*.

**Remark 5.10.** We have noted that a schema may involve the cooperative activity of multiple brain regions. In particular, then, a schema becomes a "mode of activity" for grid automata and - depending on input and context – a grid automaton can support many different schemas as their realization.

**Proposition 5.1.** If $P$ is a realization of a schema $R$, and $R$ is a concretization of a schema $Q$, then $P$ is a realization of the schema $Q$.

Corresponding to a schema $Q$ its realization R$Q$ is an operation that is also called realization.

**Corollary 5.1.** Realization of a schema is an idempotent operation.

**Definition 5.7.** A realization Rea $R$ of a schema $R$ becomes an *instantiation* of $R$ when Rea $R$ starts functioning.

Abstraction is an operation opposite to concretization.

**Definition 5.8.** An operation of changing some of the constants in a schema $P$ to variables is called an *abstraction operation* Con applied to $P$, while the result Abs $P$ of this operation is called a *abstraction* of $P$.

**Remark 5.11.** Abstraction and concretization are operations with a set of tentative results in contrast to conventional arithmetical and algebraic operations such as addition or multiplication, which give only one result (if any).

**Example 5.10.** The schema from Figure 5.1 is an abstraction of the grid automaton from Figure 2.1.

In some sense, operations of abstraction and concretization of schemas are reciprocal with respect to one another. Namely, they have the following property.

**Lemma 5.2.** a) If a schema $P$ is obtained from a schema $R$ by abstraction, then it is possible to get schema $R$ by concretization of $P$.

b) If a schema $R$ is obtained from a schema $P$ by concretization, then it is possible to get schema $P$ by abstraction of $R$.

As abstractions and concretizations are transformations of schemas, it is natural to introduce their composition as consecutive performance of corresponding transformations.

**Proposition 5.2.** a) Composition of schema concretizations is a schema concretization.
      b) Composition of schema abstractions is a schema abstraction.

**Definition 5.9.** a) A schema *P* is (*strongly*) *equivalent* to a schema *R* if they have the same realizations (concretizations).

b) A schema *P* is (*strongly*) *equivalent* to a schema *R* with respect to a class **A** of grid automata if they have the same realizations (concretizations) in **A**.

Taking an interaction schema, we are interested in its realization in the class **A** of neural networks or even more exactly, in the class **B** of neural ensembles in the brain.

**Remark 5.12.** There are other interesting equivalencies of schemas.

**Proposition 5.3.** Two schemas are strongly equivalent if and only if it is possible to obtain one from the other by renaming the variables in the first schema.

Operations of abstraction and concretization define corresponding relations in the set of all schemas.

**Definition 5.10.** If a schema *P* is obtained from a schema *R* by abstraction (or a schema *R* is obtained from a schema *P* by concretization), then *P* is called *more general* than *R* and *R* is called *more concrete* than *P*.

It is denoted by $R \geq_c P$ and $P \geq_g R$, respectively.

**Example 5.11.** The schema from Figure 5.5 is more abstract than the schema from Figure 2.1.

A *quasi-order* on a set *X* is a reflexive and transitive relation.

**Lemma 5.3**. Both $\geq_c$ and $_g\geq$ are quasiorder relations.

Concretization is a special kind of a more general operation on schemas.

**Definition 5.11.** An operation of decreasing (delimiting) the range of a variable function *f* is called a *determination operation* Det applied to *f*, while the result Det *f* of this operation is called a *determination* of *f*.

**Definition 5.12.** An operation of decreasing (delimiting) the range of the variable port assignment functions and/or port-link adjacency function of a schema *R* is called a *determination operation* Det applied to *R*, while the result Det *R* of this operation is called a *determination* of *R*.

Determination operation defines specific relations between schemas.

**Definition 5.13.** If a schema *P* is obtained from a schema *R* by determination, then *P* is called *more determined* than *R*.

It is denoted by $P \geq_{det} R$.

**Lemma 5.4**. $P_{det} \geq R$ is a relation of partial order.

It is interesting to study properties of this partial order relation for some well-known classes of schemas, e.g., program schemata.

To represent structures of schemas, we use oriented multigraphs and generalized oriented multigraphs. However, in contrast to grids of grid automata, schema grids can be not only conventional or stable oriented multigraphs and generalized oriented multigraphs, but also variable oriented multigraphs and generalized oriented multigraphs.

**Definition 5.14.** A *variable oriented* or *directed multigraph G* has the following form:

$$G = (V, E, c)$$

Here $V$ is the set of vertices or nodes of $G$; $E$ is the set of edges of $G$, each of which has the beginning and the end; the edge-node *adjacency* or *incidence function* $c: E \to V \times V$ is variable. This function assigns each edge to a pair of vertices so that the beginning of each edge is connected to the first element in the corresponding pair of vertices and the end of the same edge is connected to the second element in the same pair of vertices.

A multigraph is a graph when $c$ is an injection (Berge, 1973).

Open systems demand a more general construction.

**Definition 5.15.** A *variable generalized oriented* or *directed multigraph G* has the following form:

$$G = (V, E, c: E \to (V \times V \cup V_b \cup V_e))$$

Here $V$ is the set of vertices or nodes of $G$; $E$ is the set of edges of $G$ (with fixed beginnings and ends); $V_b \approx V_e \approx V$; the edge-node adjacency function $c$, which assigns each edge either to a pair of vertices or to one vertex, is variable. In the latter case, when the image $c(e)$ of an edge $e$ belongs to $V_b$, it means that $e$ is connected to the vertex $c(e)$ by its beginning. When the image $c(e)$ of an edge $e$ belongs to $V_e$, it means that $e$ is connected to the vertex $c(e)$ by its end. Edges that are mapped to the set $V_b \cup V_e$ are called *open*.

**Definition 5.16.** The *grid* $G(P)$ of a (basic or port) schema $P$ is the (variable) generalized oriented multigraph that has exactly the same vertices and edges as $P$, while its adjacency function $c_{G(B)}$ is $nc_B$.

**Proposition 5.5.** For any port schema $P$, we have $G(P) = G(DP)$ where $DP$ is the basic schema built from $P$.

**Example 5.12.** The grid G(*P*) of the schema *P* from Example 7 is given in Figure 5.4.

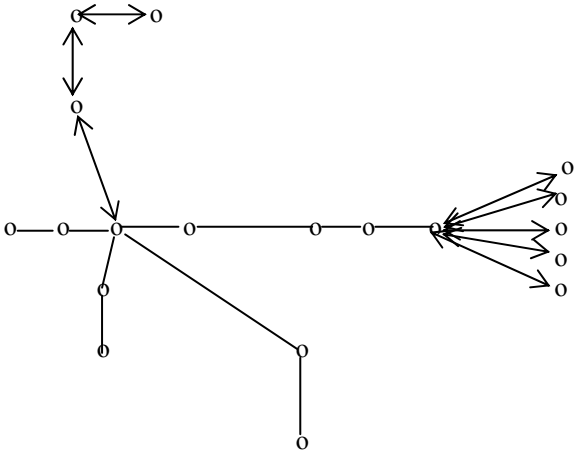

**Figure 5.4.** The grid G(*P*) of a schema

**Example 5.13.** The grid G(*P*) of the schema *P* from Example 5.12 is given in Figure 5.5.

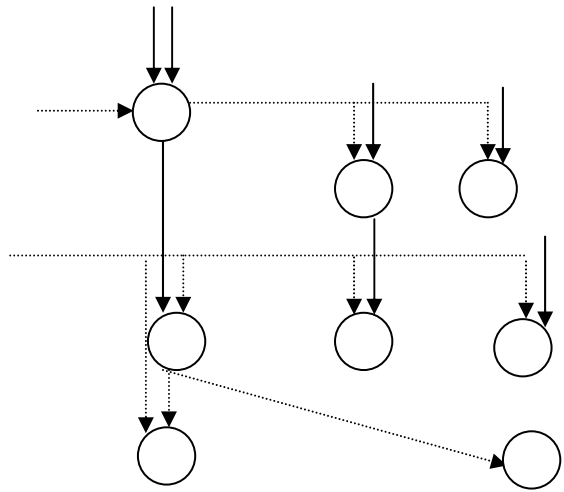

**Figure 5.5.** The grid G(*P*) of a schema

Grids of schemas allow one to characterize definite classes of schemas.

**Proposition 5.5.** A schema $B$ is closed if and only if its grid $G(B)$ satisfies the condition Im $c \subseteq V \times V$, or in other words, the grid $G(B)$ of $B$ is a conventional multigraph.

**Proposition 5.6.** A schema $B$ is an acceptor only if it has external input ports or/and its grid $G(B)$ has edges connected by their end, or Im $c \cap V_e \neq \emptyset$.

**Proposition 5.7.** A schema $B$ is a transmitter only if it has external output ports or/and its grid $G(B)$ has edges connected by their beginning, or Im $c \cap V_b \neq \emptyset$.

**Proposition 5.8.** A schema $B$ is a transducer only if it has external input and output ports or/and its grid $G(B)$ has edges connected by their beginning and edges connected by their end.

**Definition 5.17.** The *connection grid* $CG(B)$ of a port schema $B$ is the (variable) generalized oriented multigraph nodes of which bijectively correspond to the ports of $B$, while edges and the adjacency function $c_{CG(B)}$ are the same as in $B$.

**Proposition 5.9.** The grid $G(B)$ of a port schema $B$ with constant port assignment is a homomorphic image of its connection grid $CG(B)$.

Indeed, by the definition of a schema, ports are uniquely assigned to nodes, and by the definition of the grid $G(B)$ a schema $B$, the adjacency function $c_{G(B)}$ of the grid $G(B)$ is a composition of the port assignment function $p_B$ and the adjacency function $c_B$ of the schema $B$.

**Proposition 5.10**. For each schema, there is its maximal with respect to relations $_g\geq$ and $_{det}\geq$ abstraction.

Proof. It is possible to change automata, ports, and connections with variables to variables with corresponding ranges so that the new schema is equivalent to the initial one. Besides, if we have a set of variables, we can change this set to one variable with the corresponding range so that the new schema is equivalent to the initial one.

Let $R$ be a schema. We transform it in the following way. We assign to each node of its connection grid $CG(R)$ a variable the range of which encompasses all possible automata. We assign to each port of its connection grid $CG(R)$ a variable the range of which encompasses all possible ports. We assign to each edge of $G(R)$ another variable the range of which encompasses all possible links. In addition, we take as the port assignment functions and port-link adjacency function such nondeterministic functions that allow maximal flexibility of assignments and adjustments, i.e., any port may be assigned to any node or link from *R* in a permissible way and any link may be adjacent to any node or a pair of nodes.

In such a way, we obtain a maximal abstraction of *R*.

Dynamics of schemas is represented not only by operations but also by different kinds of homomorphisms.

**Definition 5.18.** A *structural homomorphism f* of a basic schema *P* into a basic schema *R* is a mapping of nodes and connections of *P* such that nodes of *P* are mapped into nodes of *R*, connections of *P* are mapped into connections of *R*, and the node-link adjacency function is preserved.

For port schemas, we have two kinds of structural homomorphisms.

**Definition 5.19.** A (*weak*) *structural homomorphism f* of a port schema *P* into a port schema *R* is a mapping of nodes and connections of *P* such that nodes of *P* are mapped into nodes of *R*, connections of *P* are mapped into connections of *R*, and the (node-link adjacency function) port assignment functions and port-link adjacency function are preserved.

It is possible to consider a structural homomorphism *f* of a basic schema *P* into a basic schema *R* as a pair of mappings: one of them $f_{(n)}$ maps nodes of *P* into nodes of *R* and the other one $f_{(e)}$ maps edges of *P* into edges of *R*. In addition, assignment functions and adjacency relation are preserved.

A structural homomorphism *f* of a port schema *P* into a port schema *R* uniquely corresponds to and determines the homomorphism $f^g$: CG(*P*) → CG(*R*) of the corresponding connection grids, while a weak structural homomorphism *f* of a port schema *P* into a port schema *R* uniquely corresponds to and determines the homomorphism $f^g$: G(*P*) → G(*R*) of the corresponding grids,.

In a natural way, compositions of structural homomorphisms and weak structural homomorphisms of port schemas as composition of mappings are introduced.

**Proposition 5.10.** Any concretization Con *P* (abstraction Abs *P*) of the schema *P* defines a structural homomorphism $f_{con}$: Con *P* → *P* ($f_{abs}$: Abs *P* → *P* ).

**Proposition 5.11.** Composition of (weak) structural homomorphisms of schemas is a (weak) structural homomorphism of schemas.

**Definition 5.20.** A (weak) structural homomorphism $f$ of a schema $P$ into a schema $R$ is called a (*weak*) *homomorphism* if the following conditions are satisfied:

a) variables from $P$ are mapped into variables and constants of the same type from $R$;

b) constants from $P$ are mapped into constants of the same type from $R$.

In a natural way, compositions of homomorphisms of schemas as composition of mappings are introduced.

**Proposition 5.12.** Composition of homomorphisms of schemas is a homomorphism of schemas.

**Proposition 5.13.** a) Any concretization Con $R$ (abstraction Abs $P$) of a schema $R$ is defined by a VE-homomorphism $c: R \rightarrow$ Con $R$ of schemas.

b) Any abstraction Abs $P$ of a schema $P$ is defined by an inverse VE-homomorphism $h: P \rightarrow$ Abs $P$ of schemas.

**Proposition 5.14.** The transformation of a schema $R$ into the corresponding basic schema D$R$ is a weak homomorphism of schemas.

Utilizing homomorphisms and structural homomorphisms, as well as Propositions 5.11 and 5.12, we build four categories of schemas: the category **SC** in which objects are schemas and morphisms are their homomorphisms; the category **GSC** in which objects are schemas and morphisms are their structural homomorphisms; the category **WSC** in which objects are schemas and morphisms are their weak homomorphisms; and the category **WGSC** in which objects are schemas and morphisms are their weak structural homomorphisms,.

**Proposition 5.15.** **SC** is a subcategory of **GSC**, while **WSC** is a subcategory of **WGSC**.

**Proposition 5.16.** **WSC** is a quotient category of **SC**, while **WGSC** is a quotient category of **GSC**.

It is possible to separate in both categories some special classes of morphisms useful in schema theory. Such morphisms represent formation and transformations of schemas.

**Definition 5.21.** a) A (structural) homomorphism of schemas $f: R \rightarrow P$ is called a (*structural*) V-*monomorphism* [E-*monomorphism*] if images of any two vertices [edges] from $R$ do not coincide. b) A (structural) homomorphism of schemas $f: R \rightarrow P$ is called a (*structural*) VE-*monomorphism* if it is both a (structural) V-monomorphism and E-monomorphism.

**Definition 5.22.** a) A (structural) homomorphism of schemas $f: R \to P$ is called a (*structural*) V-*epimorphism* [E-*epimorphism*] if any vertex [edge] from $P$ is an image of some vertex [edge] from $R$. b) A (structural) homomorphism of schemas $f: R \to P$ is called a (*structural*) VE-*epimorphism* if it is both a (structural) V-epimorphism and E-epimorphism.

**Definition 5.23.** The *image* Im $f$ of a (structural) homomorphism of schemas $f: R \to P$ is the largest subschema of $P$ such that any its vertex from $P$ is the image of some vertex [edge] from $R$ and any its edge is the image of some edge from $R$.

Let $f: R \to P$ be a (structural) homomorphism of schemas.

**Lemma 5.5.** Im $f$ is the largest subschema of $P$ such that $f$ defines an (structural) VE-epimorphism of $R$ onto Im $f$.

Let $f: R \to P$ be a (structural) E-epimorphism of schemas.

It is possible to derive properties of $R$ from properties of $P$ and vice versa.

**Proposition 5.17.** a) If the grid G($R$) is connected (full) and $f: R \to P$ is a (structural) E-epimorphism of schemas, then the grid G($P$) is also connected (full). b) If fan-in (fan-out) of all edges from the grid G($P$) is larger than $n$ and $f: R \to P$ is a (structural) E-epimorphism of schemas, then fan-in (fan-out) of all edges from the grid G($R$) is larger than $n$.

**Corollary 5.2.** a) If the grid G($P$) is disconnected and $f: R \to P$ is a (structural) E-epimorphism of schemas, then the grid G($R$) is also disconnected. b) If fan-in (fan-out) of all edges from the grid G($R$) is smaller than $n$ and $f: R \to P$ is a (structural) E-epimorphism of schemas, then fan-in (fan-out) of all edges from the grid G($P$) is smaller than $n$.

**Corollary 5.3.** If $f: R \to P$ is a (structural) E-epimorphism of schemas, then the number of components of the grid G($R$) is larger than or equal to the number of components of the grid G($P$).

**Definition 5.24.** A schema $P$ is a (*strong*) *structural subschema* of a schema $R$ if the grid $G(P)$ is a generalized oriented submultigraph of the grid $G(R)$ (the connection grid $CG(P)$ is a generalized oriented submultigraph of the connection grid $CG(R)$).

It is denoted by $P \subseteq_S R$ and $P \subseteq_{SS} R$, respectively.

**Lemma 5.6.** Any strong structural subschema of a schema $R$ is its structural subschema, i.e., $P \subseteq_{SS} R$ implies $P \subseteq_S R$.

**Example 5.14.** The schema $R$ given in Figure 5.6 is a structural subschema of the schema from Figure 5.2 and of the schema from Figure 3.1. However, the schema $R$ is neither a subschema of the schema from Figure 5.2 nor a subschema of the schema from Figure 3.1.

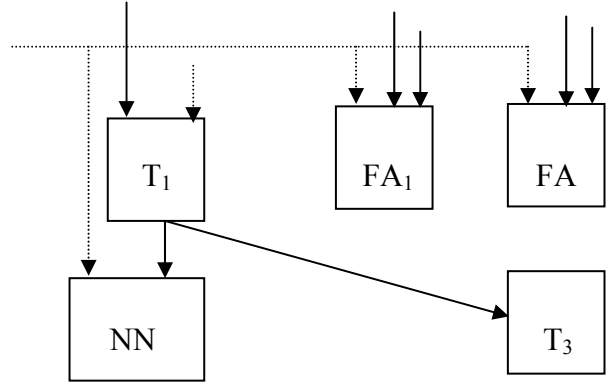

**Figure 5.6.** Dashed lines – activation of signals; solid lines – transfer of data. $T_1$ and $T_3$ are Turing machines; NN is a neural network; FA and $FA_1$ are variables the range of which is the class of all finite automata.

**Lemma 5.7.** If a schema $P$ is a *structural subschema* of a schema $R$, then there is a structural VE-monomorphism of $P$ into $R$.

**Definition 5.25.** A schema $P$ is a *subschema* of a schema $R$ if all nodes of $P$ belong to the set of nodes of $R$, all edges of $P$ belong to the set of edges of $R$, all ports of $P$ belong to the set of ports of $R$, and the internal and external port assignment functions $p_{IP}$ and $p_{EP}$ and port-link adjacency function $c_P$ of $P$ is a restriction of the internal and external port assignment functions $p_{IR}$ and $p_{ER}$ and port-link adjacency function $c_R$ of $R$, respectively. It is denoted by $P \subseteq R$.

Let *P* be a subschema of a schema *R*.

**Proposition 5.18.** For any concretization Con *P* (abstraction Abs *P*) of the schema *P*, there is a unique minimal concretization Con *R* (abstraction Abs *R*) of the schema *R* such that Con *P* (Abs *P*) is a restriction of Con *R* (Abs *R*).

**Remark 5.13.** The concept of a subschema of a schema refers to both informal and formal schemas.

**Proposition 5.19.** Any schema is a subschema of a closed schema.

In the neurophisiological schema theory (Arbib, 1995; 2005), it is important to be able to elaborate different schemas into networks of interacting subschemas until finally it becomes possible to realize these constructs in terms of neural networks or other appropriate circuitry.

**Example 5.15.** The (informal) schema of grasping given in Figure 5.7 is a subschema of the schema from Figure 3.1.

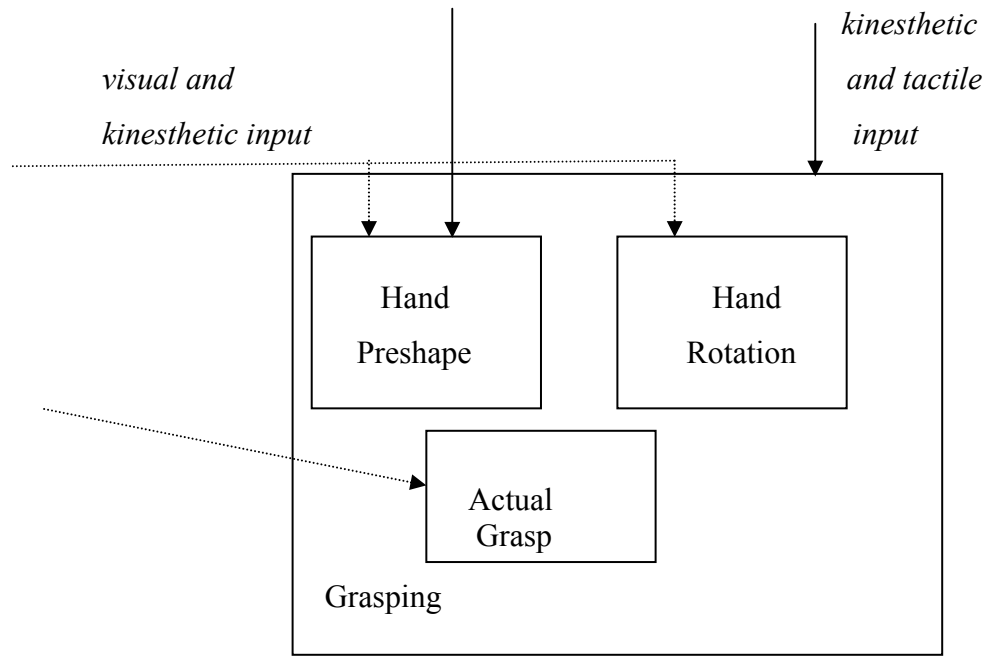

**Figure 5.7.** Dashed lines – activation of signals; solid lines – transfer of data.

**Example 5.16.** The (formal) schema given in Figure 5.8 is a subschema of the schema from Figure 5.2. This schema has connections/links of two types.

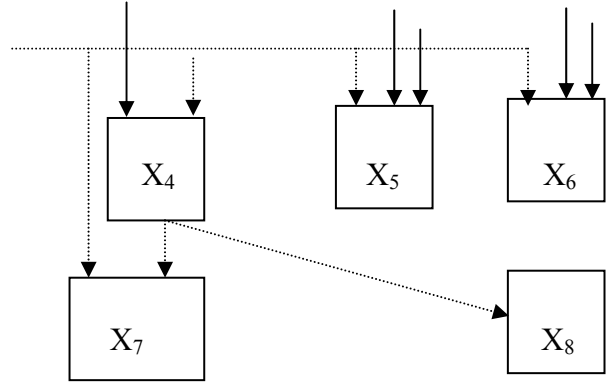

**Figure 5.8.** Dashed lines – activation of signals; solid lines – transfer of data. $X_4$ is a variable the range of which is the class of all schemas (algorithms or neural assemblages) for fast phase movement; $X_5$ is a variable the range of which is the class of all schemas (algorithms or neural assemblages) for hand preshape; $X_6$ is a variable the range of which is the class of all schemas (algorithms or neural assemblages) for hand rotation; $X_7$ is a variable the range of which is the class of all schemas (algorithms or neural assemblages) for slow phase movement; $X_8$ is a variable the range of which is the class of all schemas (algorithms or neural assemblages) for actual grasp.

**Proposition 5.20.** If $P$ is a (structural) subschema of a schema $R$, and $R$ is a (structural) subschema of a schema $Q$, then $P$ is a (structural) subschema of the schema $Q$.

**Proposition 5.21.** If $f: R \to P$ is a (structural) homomorphism [V-monomorphism, E-monomorphism, VE-monomorphism] of schemas and $Q$ is a subschema of the schema $R$, then $f$ defines a restriction $f_Q$ of $f$ on $Q$, which is a (structural) homomorphism [V-monomorphism, E-monomorphism, VE-monomorphism, respectively] of schemas

**Definition 5.26.** A subschema $Q$ of a schema $R$ is called V-complete in $R$ if $Q$ contains all nodes from $R$.

**Definition 5.27.** A subschema $Q$ of a schema $R$ is called E-complete in $R$ if $Q$ contains all edges from $R$ that are connected in $R$ to some node of $Q$.

**Definition 5.28.** A subschema $Q$ of a schema $R$ is called P-complete in $R$ if $Q$ contains all ports from $R$.

**Proposition 5.22.** A subschema $Q$ of a schema $R$ is E-complete, P-complete, and V-complete in $R$ if and only if it coincides with $R$.

**Proposition 5.23.** If a schema $R$ does not have nodes without ports, then P-completeness of a subschema $Q$ implies V-completeness of $Q$.

**Proposition 5.24.** If a schema $R$ does not have edges without ports, then P-completeness of a subschema $Q$ implies E-completeness of $Q$.

**Proposition 5.25.** If $f: R \to P$ is a structural V-epimorphism (E-epimorphism) of schemas, then the inverse image $f^{-1}(Q)$ of a V-complete (E-complete) subschema $Q$ of a schema $P$ is V-complete (E-complete).

Proof. a) Let $Q$ be a V-complete subschema of a schema $P$. Then $f^{-1}(Q)$ contains all nodes from $R$ as $f: R \to P$ is a structural V-epimorphism, i.e., $f^{-1}(Q)$ is V-complete in $R$.

b) Let $Q$ be a E-complete subschema of a schema $P$ and $r$ be an edge from $R$, at least, one end of which is connected to a node $A$ such that $f(A)$ is a node from $Q$. Then by the definition of a structural homomorphism, $f(r)$ is connected to the node $f(A)$, at least, one end. Thus, $f(r)$ is $r$ be an edge from $Q$ as $Q$ is E-complete in $R$. Consequently, $r$ belongs to $f^{-1}(Q)$. As $r$ is an arbitrary edge from $R$ connected to a node from $f^{-1}(Q)$, the schema $f^{-1}(Q)$ is E-complete in $R$.

Proposition is proved.

**Definition 5.29.** A schema $P$ is *open* if it is connected to some other systems. Otherwise, $P$ is *closed*.

Multigraphs of the schemas allow one to characterize definite classes of schemas.

**Proposition 5.25.** A schema $R$ is closed if its multigraph $G(R)$ is not generalized.

**Corollary 5.4.** A schema $R$ is a transducer only if its multigraph $G(R)$ has edges connected by their beginning and edges connected by their end.

**Proposition 5.26.** If $f: R \to P$ is a (structural) homomorphism of schemas and $R$ is a closed schema, then the image $f(R)$ is closed.

**Corollary 5.5.** An epimorhic image of a closed schema is closed.

**Proposition 5.27.** If $f: R \to P$ is a (structural) homomorphism of schemas and $R$ is a connected schema, then the image $f(R)$ is connected.

**Corollary 5.6.** An epimorhic image of a connected schema is connected.

## 5. Conclusion

A mathematical schema theory constructed in this work allows one to formalize a diversity of informal notions of schema and unify those formalized schema theories, e.g., program schema (schemata) theory, that have been used in a variety of fields. Operations on schemas and their grids are studied. Categories of schemas are constructed. All these constructions reflect situations and processes in a variety of domains.

Results obtained in this paper show how to formalize different kinds of schemas used now, how to build new classes of schemas, and how to use formalized mathematical theory to obtain properties of schemas and to study natural and artificial systems and processes by means of schemas.

The mathematical representation of schemas constructed in this work allows us to suggest the following problems for further research in the mathematical schema theory:

1. Study transformations of schemas.
2. Study compositions of schemas.
3. Study schema initiation and formation of schema assemblages.
4. Study schema functioning.
5. Study schema assemblage functioning.

## References

1. Arbib, M.A., (1985) Schemas for the Temporal Organization of Behaviour, *Human Neurobiology*, v. 4, pp.63-72

2. Arbib, M.A., *The Metaphorical Brain 2: Neural Networks and Beyond*, Wiley-Interscience, 1989

3. Arbib. M.A. Schemas and Neural Networks for Sixth Generation Computing, (1989) *Journal of Parallel and Distributed Computing*, 6:185-216.

4. Arbib, M.A., 1992, Schema Theory, in *The Encyclopedia of Artificial Intelligence*, 2nd Edition, (S.Shapiro, Ed.), Wiley-Interscience, pp. 1427-1443.

5. Arbib, M.A., 1994, Schema theory: Cooperative computation for brain theory and distributed AI, in *Artificial Intelligence and Neural Networks: Steps toward Principled Integration* (V. Honavar and L. Uhr, Eds.), Boston: Academic Press, pp.51-74.

6. Arbib, M.A., 1995, Schema Theory: From Kant to McCulloch and Beyond, in *Brain Processes, Theories and Models . An International Conference in Honor of W.S. McCulloch 25 Years After His Death*, (R. Moreno-Diaz and J. Mira-Mira, Eds.), Cambridge, MA: The MIT Press, pp.11-23.

7. Arbib, M.A., (2005) Modules, Brains and Schemas, *Formal Methods*, LNCS 3393, pp. 153-166.

8. Arbib, M. A., E.J. Conklin and J.C. Hill, *From Schema Theory to Language*, Oxford Univ. Press, 1987

9. Arbib, M.A., and Ehrig, H. (1990) Linking schemas and module specifications for distributed systems, *Proc. 2nd IEEE Workshop on Future Trends of Distributed Computing Systems*, Cairo, pp.165-171.

10. Arbib, M.A. and J.C. Hill, (1988) Language Acquisition: Schemas Replace Universal Grammar, in *Explaining Language Universals*, (J.A. Hawkins, Ed.) Basil Blackwell, pp. 56-72

11. Arbib, M.A., and Liaw, J.-S. (1995) Sensorimotor Transformations in the Worlds of Frogs and Robots, *Artificial Intelligence*, 72, pp. 53-79.

12. Arbib, M.A., Steenstrup, M., and Manes, E.G. (1983) Port Automata and the Algebra of Concurrent Processes, *Journal of Computer and System Sciences*, v. 27, pp.29-50.

13. Bartlett, F.C. *Remembering*, Cambridge University Press, 1932

14. Berge, C. *Graphs and Hypergraphs*, North Holland P.C., Amsterdam/New York, 1973

15. Bloch, A. S. *Graph-schemas and their application*, Vyshaishaya shkola, Minsk, 1975 (in Russian)

16. Blum, E.K., Ehrig, H., and Parisi-Presicce, F., 1987, Algebraic specifications of modules and their basic interconnections, *J. Comp. Syst. Sci.*, 34:293-339.

17. Burgin, M. The Block-Scheme Language as a Programming Language, *Problems of Radio-electronics*, 1973, No. 7, pp. 39-58            (in Russian)

18. Burgin, M. Recursion Operator and Representability of Functions in the Block-Scheme Language, *Programming*, 1976, No. 4, pp. 13-23 (*Programming and Computer* Software, 1976, v. 2, No.4)

19. Burgin, M. Psychological Aspects of Flow-Chart Utilization in Programming, in "*Psychological Problems of Computer Design and Utilization*," Moscow, 1985, pp. 95-96     (in Russian)

20. Burgin, M. Reflexive calculi and logic of expert systems, in *Creative processes modeling by means of knowledge bases*, Sofia, 1992, 139-160

21. Burgin, M. Flow-charts in programming: arguments pro et contra, *Control Systems and Machines*, No. 4-5, 1996, pp. 19-29     (in Russian)

22. Burgin M. (2003) From Neural networks to Grid Automata, in Proceedings of the IASTED International Conference "*Modeling and Simulation*", Palm Springs, California, pp. 307-312

23. Burgin M. (2003a) Cluster Computers and Grid Automata, in Proceedings of the ISCA 17$^{th}$ International Conference "*Computers and their Applications*", International Society for Computers and their Applications, Honolulu, Hawaii, pp. 106-109

24. Burgin, M. *Superrecursive Algorithms*, Springer, New York, 2005

25. Business Process Specification Schema (BPSS) (http://www.service-architecture.com/web-services/articles/business_process_specification_schema_bpss.html)

26. Cobas, A., and Arbib, M., 1992, Prey-Catching and Predator-Avoidance in Frog and Toad: Defining the Schemas. *J. Theor. Biol*, 157:271-304.

27. D'Andrade, R. G. 1995. *The Development of Cognitive Anthropology*. Cambridge: Cambridge University Press

28. Daum, B. *Modeling Business Objects with XML Schema*, Morgan Kauffman, 2003

29. Deloup, F. (2005) The Fundamental Group of the Circle is Trivial, *American Mathematical Monthly*, v. 112, No. 3,pp. 417-425

30. Dennis, J.B., Fossen, J.B., and Linderman, J.P. *Data Flow Schemes*, LNCS, 19, Springer, Berlin, 1974

31. Duckett, J., Ozu, N., Williams, K., Mohr, S., Cagle, K., Griffin, O., Francis Norton, F., Stokes-Rees, I., and Tennison, J. *Professional XML Schemas*, Wrox Press Ltd., 2001

32. Ehrig, H., and Mahr, B., 1985, *Fundamentals of Algebraic Specification 1: Equations and Initial Semantics*, EACTS Monographs on Theoretical Computer Science 6, Springer-Verlag.

33. Ehrig, H., and Mahr, B., 1990, *Fundamentals of Algebraic Specification 2: Module Specifications and Constraints*, EATCS Monographs on Theoretical Computer Science 21, Springer-Verlag.

34. Elmasri, R. and Navathne, S.B. *Fundamentals of Database Systems*, Addison-Wesley, Reading/Menlo Park/New York, 2000

35. Fischer, M.J. (1993) Lambda-Calculus Schemata, *LISP and Symbolic Computation*, v. 6, pp. 259-288

36. Fraenkel, A.A. and Y. Bar-Hillel, *Foundations of Set Theory*, North Holland P.C., 1958

37. Garland, S.J , and Luckham, D C (1969) *Program Schemas, Recursion Schemas and Formal Languages*, Report UCLA-EHG-7154, Los Angeles, 1971

38. Goldin, D. and Wegner, P. *Persistent Turing machines* (Brown University Technical Report, 1988)

39. Gowri, K. EnrXML – A Schema for Representing Energy Simulation Data, 7th *Int. IBPSA Conference*, Rio de Janeiro, Brazil, pp. 257-261, 2001

40. Grothendieck, A. (1957) Sur quelques points d'algebre homologique, *Tohoku Math. J.*, v. 2, No 9, pp.119-221

41. Head, H. and Holmes, G. (1911) Sensory Disturbances from Cerebral Lesions, *Brain*, v. 34, pp. 102-254

42. Heuring, V.P. and Jordan, H.F. *Computer Systems Design and Architecture*, Addison Wesley Logman, Inc., Menlo Park/Reading/Harlow, 1997

43. Herrlich, H. and Strecker, G.E. *Category Theory*, Allyn and Bacon Inc., Boston, 1973

44. Ianov, Iu I. (1958) On the equivalence and transformation of program schemes, *Communications of the ACM*, v.1 no.10, pp. 8-12

45. Ianov, Iu I. (1958a) On matrix program schemes, *Communications of the ACM*, v.1 no.12, pp. 3-6

46. Ianov, Iu I. (1958b) On the logical schemata of algorithms, *Problems of Cybernetics*, v. 1, pp. 75-127     (in Russian)

47. Itti, L and Arbib, M.A. (2005) Visual Salience Facilitates Entry into Conscious Scene Representation, in *Proc. ninth annual meeting of the Association for the Scientific Study of Consciousness (ASSC9), Pasadena, CA,*

48. Kaluznin, L.A. (1959) On Mathematical Problems Algorithmization, *Problems of Cybernetics*, v. 2, pp.                (in Russian)

49. Kant, I. 1781. *Critique of Pure Reason*, (trans. N.K. Smith) London, Macmillan, 1929

50. Karp, R.M , and Miller, R E (1969) Parallel program schemata, *J. Computer and System Science*, 3, pp. 147- 195

51. Keller, R.M. (1973) Parallel Program Schemata and Maximal Parallelism II: Construction of Closures, Journal of the ACM, v.20 no.4, pp.696-710

52. Knuth, D. *The Art of Computer Programming*, v.2: *Seminumerical Algorithms*, Third Edition, Reading, Massachusetts: Addison-Wesley, 1997

53. Knuth, D. *The Art of Computer Programming*, v.3: *Sorting and Searching*, Second Edition, Reading, Massachusetts: Addison-Wesley, 1998

54. Kotov, V.E. *Introduction to the Theory of Program Schemas*, Nauka, Novosibirsk, 1978          (in Russian)

55. Logrippo, L. (1978) Renamings and Economy of Memory in Program Schemata, Journal of the ACM, Volume 25,  no. 1 , pp. 10 - 22

56. Lyons, D.M. *RS: A Formal Model of Distributed Computation for Sensory-Based Robot Control*, in CINS Technical Report, 86-43, Department of Computer and Information Science, University of Massachusetts at Amherst, 1986

57. Lyapunov, A.A. (1958) On the logical schemata of programs, *Problems of Cybernetics*, I, pp. 46-74   (in Russian)

58. Lyons, D.M. and Arbib, M.A. (1989) A Formal Model of Computation for Sensory-Based Robotics, in *IEEE Trans. on Robotics and Automation*, v. 5, pp. 280-293

59. Margenstern, M. (2002) Cellular Automata in the Hyperbolic Plane: A Survey, *Romanian Journal of Information Science*, v. 5, No. 1/2, pp. 155-179

60. Minsky, M. (1986) *The Society of Mind*, Simon and Schuster, New York

61. Von Neumann, J. (1927) Zur Hilbertschen Bewiestheorie, *Math. Zeitschrift*, v. 26, pp. 1-46

62. Paterson, M.S., and Hewitt, C., *Comparative schematology*, MIT A.I. Lab Technical Memo No. 201 (also in Proc. of Project MAC Conference on Concurrent Systems and Parallel Computation), 1970

63. Rutledge, J. D. (1964) On Ianov's Program Schemata, *Journal of the ACM*, v. 11 , no. 1, pp. 1 - 9

64. Slutz, D. R. The flow graph schemata model of parallel computation. Rep. MAC-TRy53 (Thesis), MIT Project MAC, 1968

65. Solso, R. L. *Cognitive psychology* (4th ed.). Boston: Allyn and Bacon, 1995

66. Tsichridsis, D. and Klug, A. (Eds) *The ANSI/X3/SPARC DBMS Framework*, AFIPS Press, 1978

67. Van Der Vlist, E. *RELAX NG* , O'Reilly & Associates Incorporated, 2004

69. Vaught, R.L. (1967) Axiomatizability by a schema, *Journal of Symbolic Logic*, v. 32, no. 4, pp. 473-479

70. Weitzenfeld, A. *NSL, Neural Simulation Language, Version 1.0*, Technical Report 89-02, USC, Center for Neural Engineering, 1989

71. Weitzenfeld, A., Arbib, M.A., and Alexander, A., *The Neural Simulation Language: A System for Brain Modeling*, Cambridge, MA: The MIT Press, 2002

72. Wiedermann, J. Characterizing the Super-Turing Computing Power and Efficiency of Classical Fuzzy Turing Machines, *Theoretical Computer Science*, v. 317, No. 1/3, 2004, pp.

73. Yershov, A. P. *Introduction to Theoretical Programming*, Nauka, Moscow, 1977 (in Russian)

74. Yndurain, F.J. *Quantum Chromodynamics*, Springer-Verlag, New York/Berlin/Heidelberg, 1983

75. Zimmermann, K.-J. *Fuzzy  set theory and its applications*, Boston/Dordrecht/London, 1991